\documentclass[10pt,journal,compsoc]{IEEEtran}
\usepackage{graphicx}
\usepackage{amsmath}
\usepackage{amssymb}
\usepackage{color}
\usepackage{array}
\usepackage{ragged2e}
\ifCLASSOPTIONcompsoc
 
  \usepackage[nocompress]{cite}
\else
   \usepackage{cite}
\fi
\ifCLASSINFOpdf
\else
\fi
\begin{document}
\title{Video-based Human Action Recognition using Deep Learning: A Review}
\author{Hieu~H. Pham,
        Louahdi~Khoudour,
        Alain~Crouzil,
        Pablo~Zegers,
        and Sergio~A. Velastin
\IEEEcompsocitemizethanks{\IEEEcompsocthanksitem Hieu H. Pham and Louahdi~Khoudour are with the Centre d'Etudes et d'Expertise sur les Risques, l'environnement la mobilit\'e et l'am\'enagement (CEREMA), 1 Avenue du Colonel Roche, 31400 Toulouse, France.
\protect\\
Address all correspondence to: {\color{blue}{hieu.ph@vinuni.edu.vn}}

\IEEEcompsocthanksitem Hieu H. Pham is with the College of Engineering and Computer Science and VinUni-Illinois Smart Health Center, VinUniversity, Hanoi, Vietnam.

\IEEEcompsocthanksitem Alain~Crouzil is with the Universit\'e Paul Sabatier, Institut de Recherche en Informatique de Toulouse, 118 route de Narbonne, 31062 Toulouse Cedex 9, France.

\IEEEcompsocthanksitem Pablo~Zegers is with the Facultad de Ingenier\'ia y Ciencias Aplicadas, Universidad de los Andes, Mons. \'Alvaro del Portillo 12455, Las Condes, Santiago 7620001, Chile.

\IEEEcompsocthanksitem Sergio~A. Velastin is with the Department of Computer Science, Applied Artificial Intelligence Research Group, University Carlos III Madrid, Av. Gregorio Peces-Barba 22, Colmenarejo, 28270 Madrid, Spain. }}

\IEEEtitleabstractindextext{

\begin{abstract}
Human action recognition is an important application domain in computer vision. Its primary aim is to accurately describe human actions and their interactions from a previously unseen data sequence acquired by sensors. The ability to recognize, understand and predict complex human actions enables the construction of many important applications such as intelligent surveillance systems, human-computer interfaces, health care, security and military applications. In recent years, deep learning has been given particular attention by the computer vision community. This paper presents an overview of the current state-of-the-art in action recognition using video analysis with deep learning techniques. We present the most important deep learning models for recognizing human actions, analyze them to provide the current progress of deep learning algorithms applied to solve human action recognition problems in realistic videos highlighting their advantages and disadvantages. Based on the quantitative analysis using recognition accuracies reported in the literature, our study identifies state-of-the-art deep architectures in action recognition and then provides current trends and open problems for future works in this filed. 
\end{abstract}

}
\maketitle
\IEEEdisplaynontitleabstractindextext
\IEEEpeerreviewmaketitle

\IEEEraisesectionheading{\section{Introduction}\label{sec:introduction}}

\IEEEPARstart{I}{n} recent years, human action recognition continues to be an increasingly active research in the computer vision community due to the interest in the development of many intelligent systems involving surveillance, control, and analysis. The main goal of this area is to determine, and then predict what humans do in a video or a sequence of images. There are many potential applications such as intelligent surveillance systems \cite{ABC,Ref2,Ref3}, human-computer interfaces \cite{4383638,sonwalkar2015hand}, health care \cite{zouba2009assessing}, virtual reality \cite{7177833}, or security and military applications \cite{4270485,1217892}. 

\subsection{Motivation}

An action can be defined as a spatio-temporal sequence of human body movements.  There are many ways to define an action from the literature \cite{Moeslund200690,4633644,Poppe2010976}. Here, we consider  ``an action'' as a single motion or complex sequences of motions performed by a single person or several humans. Actions are understood as episodic examples of human dynamics that have starting and ending temporal points. From the viewpoint of computer vision, given an image sequence that contains one or many actions, human action recognition attempts to label each frame or a sequence of frames with a corresponding name of an action. In general, human action recognition is a hierarchical process, where the lower levels are on human detection and segmentation. The objective of those levels is to identify the regions of interest (ROIs) corresponding to static or moving humans in video. The visual information of actions is extracted at the next level and represented by features. These features are then used for recognizing actions. So, recognizing an action from features can be considered as a classification problem. Early attempts at human action recognition systems used independent frame-by-frame analysis methods, e.g. shape matching techniques \cite{Carlsson2001ActionRB}, while later research has focused on the spatio-temporal analysis of human motions.

A rapid increase in the number of researchers and techniques focusing on human action recognition has significantly improved its accuracy. However, action recognition is still a challenging problem due to many issues including the large intra-class difference, fuzzy boundary between classes, viewpoint, occlusion, appearance, influence of environments and recording settings \cite{Poppe2010976}, \textit{in particular from realistic videos}. Moreover, to have a complete human action recognition system, we need a mating of several disciplines including psychology and ontology \cite{Rodriguez:2014:SOH:2597757.2523819,Akdemir:2008:OBA:1459359.1459466}.

\subsection{Scope of the review, taxonomy and organization}

Human action recognition is a big topic in computer vision. Many different approaches have been published in the last two decades \cite{Aggarwal99humanmotion}. In recent years, the advances of computer vision algorithms, especially machine learning, has opened up a new direction for researchers. Therefore, it is timely that progress in this field is reviewed. In this paper, we focus on surveying publications that use deep learning, a technique that has won numerous contests in machine learning including the recognition of human actions. Our main goal is to present a review of the work that has been reported in literature, compare the performance of deep learning based approaches and other existing work in order to identify its advantages and limitations. For instance, we divide deep learning approaches for action recognition based on their architectures. Many of the most important models are covered including Convolutional neural networks (CNNs), Recurrent Neural Network with Long Short-Term memory (RNN-LSTMs), Deep Belief Networks (DBNs) and Stacked Denoising Autoencoders (SDAs). In addition, some combination architectures will also be discussed.

The review is organized as follows: First, we introduce related surveys and publicly available datasets in Section \ref{section:2}. Then, we present the key deep learning architectures for human action recognition in Section \ref{section:3}, including the main ideas and mathematical models behind each architecture. Section \ref{section:4} reviews the state-of-the-art in using deep models for human action recognition and related tasks. In Section \ref{section:5}, we give a quantitative analysis about the recognition accuracies of deep learning approaches and discuss their pros and cons. In that section, we also provide some promising directions for future research. Finally, we conclude our paper in Section \ref{section:6}.

\section{Related surveys and publicly available datasets} \label{section:2}
\subsection{Previous surveys}

In this section, we first consider related earlier surveys in human action recognition. Looking at the major conferences and journals \cite{CVPR,IVC,CVIU,MVA,SIRASU}, several earlier surveys have been published. Aggarwal and Cai \cite{aggarwal1997human} reviewed methods for human motion analysis focusing on three major areas including: motion analysis involving human body parts, tracking a moving human from a single view or multiple cameras and recognizing human activities from image sequences. Moeslund and Granum \cite{moeslund2001survey} reviewed papers on human motion capture considering a general structure for systems analyzing human body motion as a hierarchical process with four steps: initialization, tracking, pose estimation and recognition. Wang \textit{et al}. \cite{Wang_recentdevelopments} presented a survey of work on human motion analysis, in which motion analysis was illustrated as a three-level process including human detection (low-level vision), human tracking (intermediate-level vision), and behavior understanding (high-level vision). Moeslund \textit{et al}. \cite{Moeslund200690} described the work in human capture and analysis based on 280 papers from 2000 to 2006, centered on initialization of human motion, tracking, pose estimation, and recognition.

Turaga \textit{et al}. \cite{4633644} considered that ``actions" are characterized by simple motion patterns typically executed by a single person while ``activities" are more complex and involve coordinated actions among a small number of humans and reviewed the major approaches for recognizing human action and activities. Poppe \cite{Poppe2010976} focused on image representation and action classification methods. A similar survey by Weinland \textit{et al}. \cite{Weinland2011224} also concentrated on approaches for action representation and classification. Popoola and Wang \cite{6129539} presented a survey focusing on contextual abnormal human behavior detection for surveillance applications. Ke \textit{et al}. \cite{ke2013review} reviewed human activity recognition methods for both static and moving cameras, covering many problems such as feature extraction, representation techniques, activity detection and classification. Aggarwal and Xia \cite{Aggarwal201470} presented a survey of human activity recognition based on 3D data, especially on using RGB and depth information acquired by consumer 3D sensors as the Kinect \cite{zhang2012microsoft} sensor. Guo and Lai \cite{guo2014survey} gave a survey of existing approaches on still image-based action recognition.

Recently, Cheng \textit{et al}. \cite{cheng2015advances} reviewed approaches on human action recognition using an approach-based taxonomy, in which all methodologies are classified into two categories: single-layered approaches and hierarchical approaches. In addition, Vrigkas \textit{et al}. \cite{10.3389/frobt.2015.00028} categorized human activity recognition methods into two main categories including ``unimodal" and ``multimodal". Then, they reviewed classification methods for each of these two categories. The survey of Subetha and Chitrakala \cite{7518920} mainly focused on human activity recognition and human-object interaction methods. Presti \textit{et al.} \cite{LoPresti:2016:SHA:2894384.2894499} provided a survey of human action recognition based on 3D skeletons, summarizing the main technologies, including both hardware and software for solving the problem of action classification inferred from  time series of 3D skeletons. In addition, another survey was presented by Kang and Wildes \cite{Kang2016ReviewOA}. It summarized various action recognition and detection algorithms, focused on encoding and classifying features. The latest survey on human action recognition was published in early 2016 by Herath  \textit{et al.} \cite{DBLP:journals/corr/HerathHP16}, in which the authors reviewed methods based on hand-crafted features and some deep architectures for recognizing actions. Table~\ref{tab1} summarizes previous surveys on human action and activity recognition published from 1997 to 2017 and reviewed in this paper. The surveys in the literature have shown that the common approaches in human action recognition have focused on using hand-designed local features such as HOG/HOF \cite{dalal2005histograms,laptev2008learning}, SIFT \cite{lowe1999object}, or SURF \cite{bay2006surf}. In addition, these approaches are also extended for more robustness in video processing such as Cuboids \cite{dollar2005behavior}, HOG3D \cite{klaser2008spatio}. To the best of our knowledge, there is no review on human action recognition based on deep learning techniques including comparisons of the performance of deep learning based approaches with traditional methods and with each other. Moreover, deep learning is a rapidly growing field, where novel algorithms appear in very short time duration and change the way of understanding and recognizing actions from visual data. That has prompted us to perform this work. Not only to provide a comparative analysis about the current state of human action recognition using deep learning algorithms, but also to point out the new trends in this field. Our survey will add to the latest reviews on human action recognition in the literature.

\begin{table}[h!]  
  \centering
  \caption{\textbf{Summary of previous surveys and their key points ordered by year of publication.}} 
  \label{tab1}
  \begin{tabular}{p{0.3\linewidth}p{0.05\linewidth}p{0.5\linewidth}}
    \hline
    \textbf{Authors} & \textbf{Year} &  \textbf{Main topics / Area of Interest}\\
    \hline
    Aggarwal \textit{et al.} \cite{aggarwal1997human} 		& 1997 &  Human motion analysis, tracking.\\
    Moeslund \textit{et al.} \cite{moeslund2001survey} 	& 2001 &  Motion initialization, tracking, pose estimation, recognition.\\										
    Wang \textit{et al}. \cite{Wang_recentdevelopments} & 2003 & Human detection, tracking, activity understanding.\\  													
    Moeslund \textit{et al}. \cite{Moeslund200690} & 2006 & Human motion capture, action, and behavior analysis.\\
    Turaga \textit{et al}. \cite{4633644} 		   & 2008 & Recognizing human behavior.\\
    Poppe \cite{Poppe2010976} 					   & 2010 & Feature extraction and classification of human action.\\
    Weinland \cite{Weinland2011224} 			   & 2011 & Full-body action segmentation, and recognition.\\ 													
    Popoola \textit{et al.} \cite{6129539} 			   & 2012 &  Human motion analysis, abnormal behavior recognition.\\
    Ke \textit{et al}. \cite{ke2013review} 		   & 2013 & Human activity recognition from static and moving camera.\\
    Aggarwal \textit{et al.} \cite{Aggarwal201470} 		   & 2014 & Human activity recognition from 3D and depth data.\\
    Guo \textit{et al.} \cite{guo2014survey} 			   & 2014 & Human action recognition using still image.\\ 
    Cheng \textit{et al}. \cite{cheng2015advances} & 2015 & Single-layered and hierarchical approaches for action recognition.\\
    Vrigkas \textit{et al}. \cite{10.3389/frobt.2015.00028} & 2015 & Human activity classification.\\
    Subetha \textit{et al.} \cite{7518920} 		   & 2016 & Human activity recognition and human-object interactions. \\
    Presti \textit{et al.} \cite{LoPresti:2016:SHA:2894384.2894499} & 2016 & Action classification based on skeleton. \\
    Kang \textit{et al.} \cite{Kang2016ReviewOA} & 2016 & Human action recognition and detection.\\	
    Herath  \textit{et al.}  \cite{DBLP:journals/corr/HerathHP16} & 2016 & Human action recognition based on handcrafted features and deep learning approaches \\
    
	\hline
  \end{tabular} 
\end{table}		
\subsection{Benchmark datasets for human action recognition}

With the increase in study of human action recognition algorithms, many datasets have been recorded and published for the research community. Much of the progress in action recognition was demonstrated on standard benchmark datasets. These datasets allow us to develop, evaluate and compare new methods. In this section we summarize the most important public datasets in the area. From the early dataset which contained very simple actions and acquired under controlled environments, to recent benchmark datasets with thousands of video samples and millions of frames providing complex actions and human behaviors from the real world. Table~\ref{tab2} shows the datasets and their descriptions. To guide readers in the selection of the most suitable dataset for evaluating their work, we divide benchmarks into four categories including single action (category I), human-human interaction, human-object interaction and behavior (category II), surveillance (category III) and sport videos and other types (category IV).
\begin{table*}
  \centering
  \caption{\textbf{Some popular datasets for human action recognition (ordered by year of publication).}} 
  \label{tab2} 
  \begin{tabular}{ccccc} 
    \hline 
   \textbf{ Dataset (category)} & \textbf{Author} & \textbf{ Year} & \textbf{\# Classes} & \textbf{Description} \\
    \hline 
    {\small KTH} & {\small Schuldt \textit{et al.} \cite{schuldt2004recognizing}} & 2004  & {\small 6} & 
    {\small Walking, jogging, running, boxing,}\\
    {\small (I)} & & &  &
    {\small hand waving, and hand clapping.}\\
    \\
   {\small Weizman} & {\small Gorelick \textit{et al.} \cite{ActionsAsSpaceTimeShapes_pami07}} & 2005 &{\small 10} & 
   {\small Walk, run, jump, gallop sideways,}
   \\
   {\small (I)} &  & &  &
   {\small bend, one-hand wave, two-hands wave,}
   \\
   & & &  &
   {\small jump in place, jumping, jack skip.}
   \\
   \\
    {\small IXMAS} & {\small Weinland \textit{et al.} \cite{weinland06}} &  2006 &{\small 13} & 
   {\small Check watch, cross arms, scratch head,}
   \\
   {\small (I)} & & & &
   {\small sit down, get up, turn around, walk,}
   \\
   & & &  &
   {\small wave, punch, kick, point, pick up, etc.} \\
	\\
	{\small Hollywood-1} & {\small Laptev \textit{et al.} \cite{4587756}} & 2008 &{\small 8} & {\small Answer phone, get out car, hand shake,hug } \\
	{\small (II)} & & & & {\small person, kiss, sit down, sit up, stand up.}\\
	\\
    {\small Hollywood-2} & {\small Marszalek \textit{et al.} \cite{5206557}} & 2008 &{\small 12} & {\small Answer phone, drive car, eat, } \\
	{\small (II)} &  & & & {\small fight person,hug person, kiss, run,etc.}\\
	\\    
	{\small YouTube} & {\small Liu \textit{et al.} \cite{liu2009recognizing}} & 2009 & {\small 8} & {\small Basketball shooting, volleyball spiking,} \\
	 {\small (II)} &  & & &{\small soccer juggling, cycling, diving, etc.}\\
	\\
  	{\small MuHAVi} & {\small Singh \textit{et al.} \cite{singh2010muhavi}} & 2010  & {\small 17} & {\small Walk turn back, run stop, punch, kick,}\\
   	{\small (II)} &  & & &{\small shot gun collapse, pull heavy object,}\\
   	& & & &{\small pick up through object, walk fall.}\\
   	\\
   	{\small UT-Interaction} & {\small Ryoo \textit{et al.} \cite{UT-Interaction-Data}} & 2010 &{\small 6} & {\small Shake-hands, point, hug, push, kick,}\\
   {\small (II)} &  & & &{\small and punch.} \\
   \\
	{\small MSR Action3D} & {\small Li \textit{et al.} \cite{li2010action}} & 2010  &{\small 20} & {\small High arm wave, horizontal
	arm wave, hammer, } \\
	{\small (II)} & & & &{\small hand catch, forward punch, high throw, etc. }\\
	\\
	{\small Daily-Activity-3D} & {\small Wang \textit{et al.} \cite{wang2012mining} } &  2010 & {\small 16} & {\small Drink, eat, read book, call cellphone, } \\
	{\small (II)} & & & &{\small cheer up, sit still, toss paper, play game, etc. }\\
	\\
     {\small MSR Action3D} & {\small Li \textit{et al.} \cite{Li2010ActionRB}} & 2010 &{\small 20} & 
   {\small High arm wave, horizontal arm wave, hammer,}
   \\
   {\small (II)} &  & &  &
   {\small hand catch, forward punch, high throw, etc} \\
   \\
	{\small Olympic Sports} & {\small Niebles \textit{et al.} \cite{Niebles2010}} & 2010 &{\small 16} & {\small High jump, long jump, triple jump, pole vault} \\
    {\small (IV)} &  & & &{\small discus throw, hammer throw, etc.}  \\
    \\
    {\small VIRAT 2.0} & {\small Oh \textit{et al.} \cite{5995586}} & 2011 &{\small 12} & {\small Loading an object to a vehicle, opening a} \\
	{\small (III)} &  & &  &{\small vehicle trunk, getting into a vehicle, etc.}\\
	\\
    {\small HMDB-51 } & {\small Kuehne \textit{et al.}  \cite{6126543}} &  2011 & {\small 51} & {\small Smile, laugh, chew, talk,} \\
    {\small (II)} & & & & {\small smoke, eat, drink, etc.}  \\
    \\
	{\small Cornell Activity CAD-60 } & {\small Sung \textit{et al.} \cite{Sung2011HumanAD}} &  2011 & {\small 12} & {\small Rinsing mouth, brushing teeth,} \\
	
	{\small (II)} & & & &{\small talking on the phone, drinking water, etc. }\\
	\\
    {\small Cornell Activity CAD-120 } & {\small Koppula \textit{et al.} \cite{DBLP:journals/corr/abs-1210-1207}} & 2012 & {\small 20} & {\small Making cereal, taking medicine, stacking objects,} \\
	
	{\small (II)} &  & &  &{\small reaching, moving, pouring, eating, etc. }\\
	\\
	{\small SBU-Kinect Interaction} & {\small Kiwon \textit{et al.} \cite{kiwon_hau3d12}} & 2012 &{\small 8} & {\small Approach, depart, push, kick,  punch,}\\ (II) &  & & &{\small exchange objects, hug, and shake hands.}\\
   \\
    {\small LIRIS} & {\small Wolf \textit{et al.} \cite{wolf2014evaluation}} & 2012 & {\small 10} & 
   {\small Discussion between two or more people,}
  \\
   {\small (II)} &  & &  &
   {\small give an object to another person,}
   \\
   & & &  &
   {\small put (take) an object into (from) a box (desk), etc.}\\
   \\
    {\small UCF-50 } & {\small Reddy \textit{et al.}  \cite{Reddy2013}} &  2012 & {\small 50} & {\small Diving, drumming, fencing, } \\
    {\small (IV)} & & & &{\small tennis swing, trampoline jumping, playing piano, etc.}  \\
    \\
    {\small UCF-101 } & {\small Soomro \textit{et al.}  \cite{DBLP:journals/corr/abs-1212-0402}} & 2012 &{\small 101} & {\small Horse riding, hula hoop, ice dancing, } \\
    {\small (IV)} &  & & & {\small skiing, skijet, sky diving, etc.}  \\
    \\
    {\small Sports-1M } & {\small  Karpathy \textit{et al.} \textit{et al.}  \cite{Karpathy2014LargeScaleVC}} &  2014 & {\small 487} & {\small Juggling club, pole climbing, tricking, } \\
    {\small (IV)} & & & & {\small foot-bag, skipping rope, slack-lining, etc.} \\
    \\
    {\small ActivityNet } & {\small Heilbron  \textit{et al.}  \cite{caba2015activitynet}} & 2015 & {\small 203} & {\small Personal care, eating and drinking, household,} \\
    {\small (II)} &  & & & {\small caring and helping, working, socializing, etc.}  \\
    \\
    {\small NTU RGB+D } & {\small Shahroudy  \textit{et al.}  \cite{DBLP:journals/corr/ShahroudyLNW16}} & 2016 & {\small 60} & {\small Drinking, eating, reading, } \\
    {\small (II)} &  & & & {\small punching, kicking, hugging, etc.}  \\
    \\     
	\hline
  \end{tabular}
\end{table*} 
The complexity of each dataset depends on its recorded setting. For example, early benchmark datasets such such as KTH \cite{schuldt2004recognizing} or Weizmann \cite{ActionsAsSpaceTimeShapes_pami07} were made under laboratory conditions for idealized human actions: all of them are composed of simple and unrealistic actions and homogeneous background. Many methods have already achieved very high recognition rates on these datasets. Performances have increased over the years and have reached perfect accuracy, e.g., 100\% on the Weizman \cite{ActionsAsSpaceTimeShapes_pami07} by Ikizler \textit{et al.} \cite{Ikizler2007HumanAR} or Brahnam \textit{et al.} \cite{Brahnam2009HighPS}. In other words, we can say that the unrealistic datasets have already already solved by our action recondition systems. Another dataset named IXMAS has also been produced under laboratory conditions, but with multiple viewpoints \cite{weinland06}.

\begin{figure}
\begin{center}
\begin{tabular}{c}
\includegraphics[width=8.9cm,height=8.9cm]{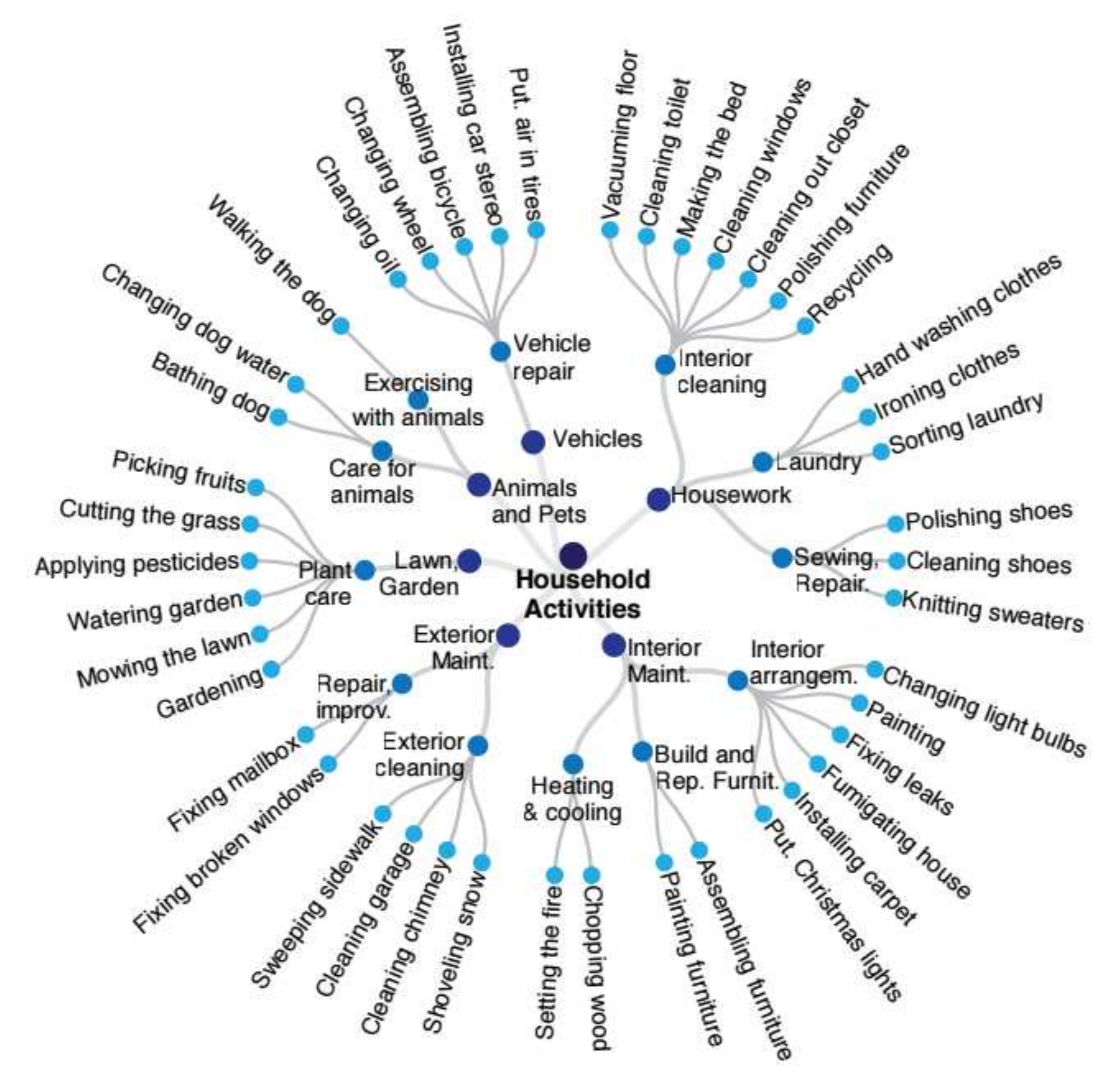}  
\end{tabular}
\end{center}
\caption 
{ \label{figntu4} Household activities from the ActivityNet \cite{caba2015activitynet} dataset.}
\end{figure}

\begin{figure}
\begin{center}
\begin{tabular}{c}
\includegraphics[width=8.9cm,height=6cm]{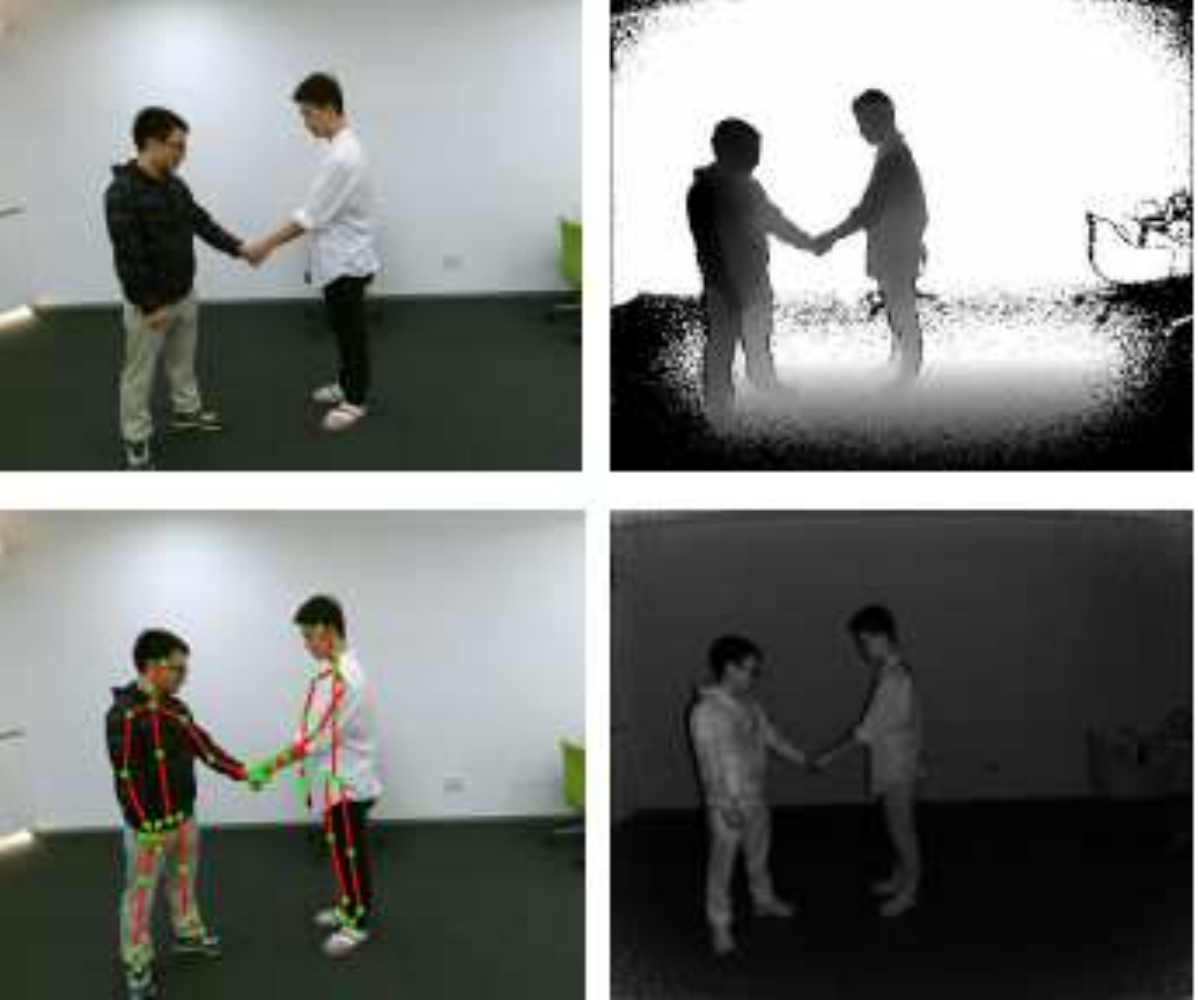}  
\end{tabular}
\end{center}
\caption 
{ \label{figntu} Some samples of RGB, depth, skeleton and IR image from the NTU RGB+D dataset \cite{DBLP:journals/corr/ShahroudyLNW16}.}
\end{figure}

After the success of the action recognition systems on benchmarks produced ''in the lab'', more complex benchmarks have been released. For instance, MSR Action3D \cite{li2010action}, UT-Interaction \cite{UT-Interaction-Data}, Daily-Activity-3D \cite{wang2012mining}, Cornell Activity CAD-60 \cite{CAD_60_database}, Cornell Activity CAD-120 \cite{DBLP:journals/corr/abs-1210-1207}, VIRAT 2.0 \cite{5995586}, SBU-Kinect Interaction \cite{kiwon_hau3d12}. These datasets aim to provide challenging videos of human action under unconstrained environments with complex background and illumination conditions. However, they are not ''real'' actions. Then, many researchers have extracted realistic situations from movie or sport videos on social networks such as YouTube to make new realistic benchmark datasets. See for example Hollywood-1 \cite{4587756}, Hollywood-2 \cite{5206557}, YouTube \cite{liu2009recognizing},  HMDB-51 \cite{6126543}, UCF-50 \cite{Reddy2013}, UCF-101 \cite{DBLP:journals/corr/abs-1212-0402}, Sports-1M \cite{Karpathy2014LargeScaleVC}, ActivityNet \cite{caba2015activitynet}. The general approach in these datasets is to collect videos from ``in-the-wild'' sources with many clips and action classes. It is easy to see that several datasets are designed with deep learning algorithms in mind due to their very large scale. For example, in Sports-1M \cite{Karpathy2014LargeScaleVC} there are around 1 million YouTube videos belonging to a taxonomy of 487 classes of sports, ActivityNet \cite{caba2015activitynet} provides more than 200 activity classes with 10,024 training videos, 4,926 validation videos and 5,044 testing videos. Figure~\ref{figntu4} shows some actions in a class of the ActivityNet \cite{caba2015activitynet} dataset. These large scale datasets are an important premise for the development of deep learning methods because they require a large number of training data and tuning them on small and out-of-date datasets such as KTH \cite{schuldt2004recognizing} or Weizmann \cite{ActionsAsSpaceTimeShapes_pami07} leads to low performance.
Most recently, Shahroudy \textit{et al.} introduced NTU RGB+D dataset \cite{DBLP:journals/corr/ShahroudyLNW16}, a very large-scale RBD-D dataset for human action recognition. The NTU RGB+D dataset contains more than 56 thousand video samples, 4 million frames with 60 different action classes and performed by 40 different subjects. To our best knowledge, this is the newest dataset for action recognition tasks. Some samples of RGB, depth, joints, and IR image are shown in Figure~\ref{figntu}.
Experiments on realistic human action datasets have so far given limited results specially when dealing with a large and varied range of actions (e.g., table~\ref{tab3} shows recognition results methods on the HMDB-51 \cite{6126543} dataset). Therefore, the current problem in action recognition that needs solving by computer vision community is recognizing \textit{complex actions and behaviors} on \textit{realistic scenarios}. Furthermore, there is also the need to build cost-effective real-world applications. This explains state-of-the-art benchmark datasets such as UCF-101 \cite{DBLP:journals/corr/abs-1212-0402}, HMDB-51 \cite{6126543}, Sports-1M \cite{Karpathy2014LargeScaleVC}, ActivityNet \cite{caba2015activitynet} and NTU RGB+D \cite{DBLP:journals/corr/ShahroudyLNW16}. Researchers who want to evaluate their algorithms on state-of-the-art benchmark datasets can participate in the THUMOS challenge \cite{Idrees2017TheTC}, a common benchmark for action classification and detection for computer vision community from around the world.   

Recent developments in low-cost depth sensor technology have brought many opportunities for solving human action recognition tasks. RGB-D and skeleton data allow better understanding of the 3D structure of human body motion. Related to RGB-D and skeleton datasets, interested readers are referred to the recent work of Zhang \textit{et al.} \cite{Zhang2016RGBDbasedAR} and Firman \cite{DBLP:journals/corr/Firman16}. In the next section, we will present deep learning-based approaches, one of the most interesting techniques in recent years in this field to answer the challenges highlighted here.
\begin{table}[h!]
  \centering
  \caption{\textbf{Accuracy on the HMDB-51 dataset \cite{6126543}}}
  \label{tab3}
  \begin{tabular}{cccc}
  \hline
  Approach    & Author & Year & Acc.(\%)\\ 
    \hline
    RGB + optical flow fusion &  Wang \textit{et al.} \cite{Wang2016ActionsT} & 2016 & \textbf{62.0}\\
    $F_{ST}$ + SCI fusion & Sun \textit{et al.} \cite{Sun2015HumanAR} & 2015 & 59.1\\
    Two-stream CNN + SVM  & Simonyan \textit{et al.} \cite{simonyan2014two} & 2014 & 59.4\\
   Improved dense trajectory & Wang \textit{et al.}  \cite{wang2013action} & 2013 & 57.2\\ 
   W-flow dense trajectories & Jain\textit{et al.}  \cite{jain2013better} & 2013 & 52.1\\ 
   Dense trajectory & Wang \textit{et al.}  \cite{wang2013dense} & 2013 & 46.6\\ 
   TRAJMF & Jiang \textit{et al.}  \cite{jiang2012trajectory} & 2012 & 40.7\\ 
   Binary ranking models & Can \textit{et al.}  \cite{can2013formulating} & 2013 & 39.0\\ 
   MIP & Kliper-Gross \textit{et al.}  \cite{KliperGross2012MotionIP} & 2012 & 29.2\\ 
   GIST 3D & Solmaz \textit{et al.}  \cite{Solmaz2012ClassifyingWV} & 2012 & 	29.2 \\ 
   Action bank & Sadanand \textit{et al.}  \cite{Sadanand2012ActionBA} & 2012 & 26.9 \\ 
   C2 & Kuehne \textit{et al.}  \cite{6126543} & 2011 & 23.0\\ 
   HOG/HOF & Kuehne \textit{et al.}  \cite{6126543} & 2011 & 20.0\\  
  	\hline 
  \end{tabular}
\end{table}

\section{Deep Learning: a short presentation} \label{section:3}

For the sake of completeness, we present this section especially for readers who might not be very familiar with deep learning techniques. A full discussion is clearly outside the scope of this paper. Before discussing deep learning, we would like to briefly summarize the concept of machine learning (ML). ML is the branch of algorithms that allows computers to automatically learn from data. We can use ML systems for identifying objects in images, detecting spam emails, understanding text, finding genes associated with a particular disease and numerous other applications. The primary goal of ML is to develop general-purpose algorithms which are able to make accurate predictions in many different tasks.\textit{ In other words, ML algorithms try to match the density function that produced the data}. For example in classification problems, we need to identify a set of categories $ \mathcal{C} $ from a space of all possible examples $\mathcal{X}$. Given any set of labeled examples $ (\textbf{x}_{1}, \textbf{c}_{1}), . . . ,(\textbf{x}_{m}, \textbf{c}_{m})$ , where $ \textbf{x}_{i} \in \mathcal{X} $ and $ \textbf{c}_{i} \in \mathcal{C} $; the goal of ML is to find a concept $ \mathcal{F}(\cdot)$ that satisfies $ \textbf{c}_{i} = \mathcal{F}(\textbf{x}_{i}) $ for all $i$.  In general, ML algorithms include two main steps. The first step is to define the representations of the raw data acquired by sensors, called ``\textit{feature extraction}". Then, these features are mapped into labels and called ``\textit{feature-to-label mapping}". This process produces an ML model that can be applied for new unlabeled data. Depending on the way of learning, (e.g., learn from labeled data or unlabeled data, learn with feedback or non-feedback), ML methods are typically classified into four categories including supervised learning, unsupervised learning, semi-supervised learning, reinforcement learning.

Deep Learning (DL) is a class of techniques in machine learning. In 2012, DL became a major breakthrough in computer vision after the authors of AlexNet \cite{NIPS2012_4824} achieved record performance on a highly challenging dataset named ImageNet. AlexNet \cite{NIPS2012_4824} was able to classify 1.2 million high-resolution images from 1000 different classes with the best error rate. In general, DL methods are machine learning methods that consists and operate on multiple (multi-layer) levels  of representation.

Various DL architectures have been proposed over the years (see Table ~\ref{tab4}) and have been shown to produce state-of-the-art results on many tasks, not least within human action recognition. In this section, we describe the most important DL architectures for human action recognition including Convolutional Neural Networks (CNNs or ConvNets) \cite{Fukushima1980,rumelhart1988learning,lecun1989backpropagation,Krizhevsky_imagenetclassification}, Recurrent Neural Networks with Long Short-Term Memory (RNN-LSTMs) \cite{hochreiter1997long}, Deep Belief Networks (DBNs) \cite{hinton2006fast}, and Stacked Denoising Autoencoders (SDAs) \cite{vincent2008extracting}.
\begin{table}[h!]
  \centering
  \caption{\textbf{Popular deep learning architectures}.}
  \label{tab4}
  \begin{tabular}{p{0.3\linewidth}p{0.5\linewidth}}
  \hline
  Architecture    & Main articles \\ 
    \hline
    CNNs & Fukushima (1980) \cite{Fukushima1980}; \\
    	 & Rumelhart \textit{et al.} (1986) \cite{rumelhart1988learning}; \\
    	 & LeCun \textit{et al.} (1989) \cite{lecun1989backpropagation}; \\
    	 & Krizhevsky \textit{et al.} (2012) \cite{Krizhevsky_imagenetclassification} \\
         & Szegedy \textit{et al.} (2015) \cite{Szegedy2015GoingDW} \\
         & Simonyan \textit{et al.} (2014) \cite{Simonyan2014VeryDC} \\
         & He \textit{et al.} (2015) \cite{he2015deep} \\
         
    RNN-LSTMs & Hochreiter and Schmidhuber \cite{hochreiter1997long} \\
         
  	DBNs & Hinton et al. \cite{hinton2006fast} \\
    
  	DBMs & Salakhutdinov et al. (2006)\cite{salakhutdinov2009deep} \\
    
  	Sparse Coding 	& Olshausen and Field (1996)\cite{olshausen1996emergence}; \\
  									& Lee et al. (2006) \cite{lee2006efficient} \\
                                    
  	SDAs			& Vincent et al. (2008) \cite{vincent2008extracting} \\
  	\hline
  \end{tabular}
\end{table}
\subsection{Convolutional neural networks (CNNs)}
\label{label:cnn}
\begin{figure*}
\begin{center}
\begin{tabular}{c}
\includegraphics[width=15cm,height=5cm]{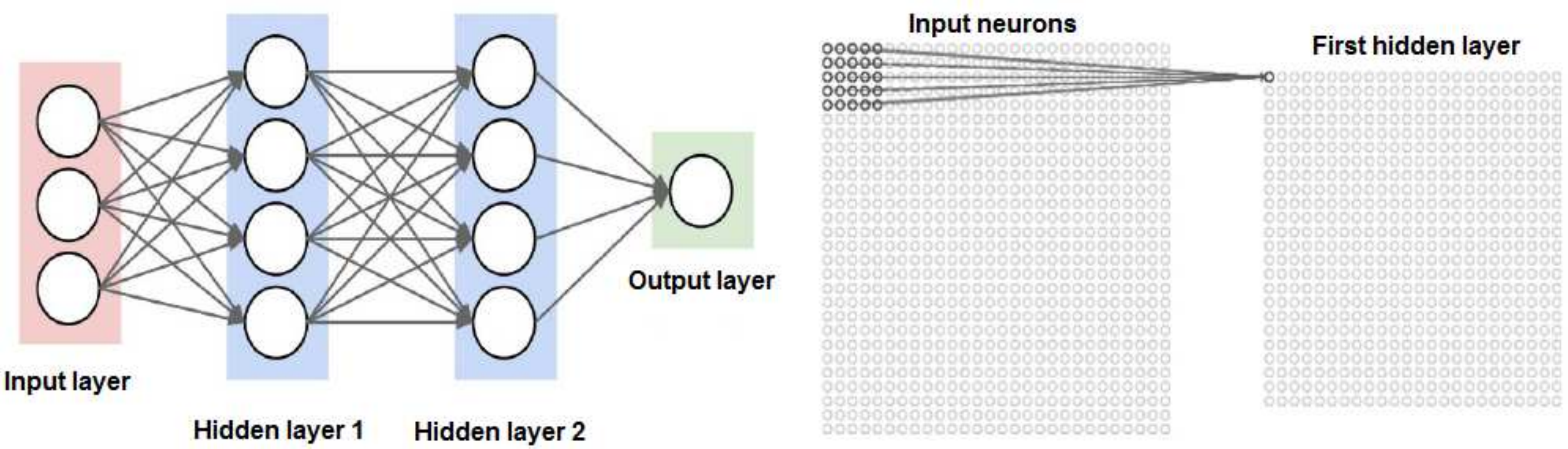}\\
\hspace*{0.001cm} (a) \hspace{7.3cm} (b) 
\end{tabular}
\end{center}
\caption 
{ \label{fig6} \textbf{(a)} Illustration of a fully-connected model in a regular 3-layer neural network. \textbf{(b)} Illustration of the local receptive field in the input neurons \cite{neuralnetworksanddeeplearning}.} 
\end{figure*}

After obtaining breakthrough results in object recognition with AlexNet \cite{NIPS2012_4824} for the ImageNet project in 2012, CNNs become one of the most important deep learning models and play a dominant role for solving visual-related tasks. A CNN is a type of artificial neural network, designed for processing visual and other two-dimensional data. The main benefit of this model is that it operates directly on the raw data without any hand-crafted feature extraction. The idea of CNNs was firstly presented in 1980 by Fukushima \cite{Fukushima1980} inspired by the structure of the visual nervous system \cite{hubel1962receptive}. CNN models continued to be proposed and developed e.g. by Rumelhart \textit{et al}. \cite{rumelhart1988learning}, LeCun \textit{et al}. \cite{lecun1989backpropagation} and Krizhevsky \textit{et al}. \cite{Krizhevsky_imagenetclassification}. There are three key ideas behind a CNN architecture including \textit{''local connections''}, \textit{''shared weights''}, and \textit{''pooling''}. \\
\\
\textbf{Local connections}: In regular neural networks, each hidden layer consists of a set of neurons, where each neuron is fully connected to all neurons in the previous layer (Figure~\ref{fig6}a). This model does not work efficiently when the input vector has a hight dimension. To make this more efficient, the idea is to reduce the number of connections between the first hidden layer to the input or each hidden layers to each other. Given an image as an input vector, every input pixel is not connected to every neuron in the first hidden layer. Instead, neurons in the first hidden layer are connected to localized regions of the input image. This sub-region is called the ``\textit{local receptive field}". For each local receptive field, we can identify a neuron in the first hidden layer as shown in Figure~\ref{fig6}b. \\
\\
\textbf{Shared weights}: For standard neural networks such as multilayer perceptrons \cite{ruck1990feature} (MLP), the neurons of the first layer are computed by the dot product function of input vector $\vec{\textbf{x}}$ and its weights $\vec{\textbf{w}}$ where many different $\textbf{w}_{i}$ values are used. In a CNN, we use a technique called ``\textit{weight sharing}" which is able to reduce the number of parameters $\textbf{w}_{i}$. Specifically, in weight sharing, some of the parameters in the CNN model are constrained to be equal to each other \cite{levietquoctutorial}. Mathematically, the weight sharing technique can be performed using a convolution operator. In this process, we apply the filters to many local receptive fields in the input image, a ``\textit{feature map}" is generated by sliding a filter over the input matrix and computing the dot product. We can use many different filters and each of them will produce one feature map.\\
\\
\textbf{Pooling}: ``\textit{Pooling}" also called ``\textit{subsampling}" is a sample-based discretization process. Its main goal is to reduce the dimensionality of the input representation while retaining the most important information in feature maps. This process reduces the computational cost and at the same time it provides a form of translation invariance. Max-pooling is performed by applying a max filter, it computes the max value of a selected set of output neurons from the feature map in the convolutional layer.\\
\\
These concepts above can now be put together to form a complete CNN architecture that consists of a series of stages. The first few stages are structured by one convolutional layer and one max-pooling layer. These layers are followed by one or more fully connected layers at the top of the model. In a CNN, the convolution layer plays the role of a local feature extractor while the max-pooling layer  merges semantically similar  features into one. The last layer is a standard neural network working as a classifier (or a standard classifier such as an SVM). So the network learns a set of good features (c.f. with arbitrarily chosen or hand-crafted features) to use with a classifier. To prevent over-fitting and train the CNNs faster, Rectified Linear Units (ReLUs) and Dropout Layers \cite{Srivastava2014DropoutAS,Dahl2013ImprovingDN} have also been used. However, we do not discuss these two layers here as it is beyond the scope of this paper.

\subsection{Recurrent Neural Networks with Long-Short Term Memory (RNN-LSTMs)}
\label{subsection:3.4}

\begin{figure*}
\begin{center}
\begin{tabular}{c}
\includegraphics[width=12cm,height=3.5cm]{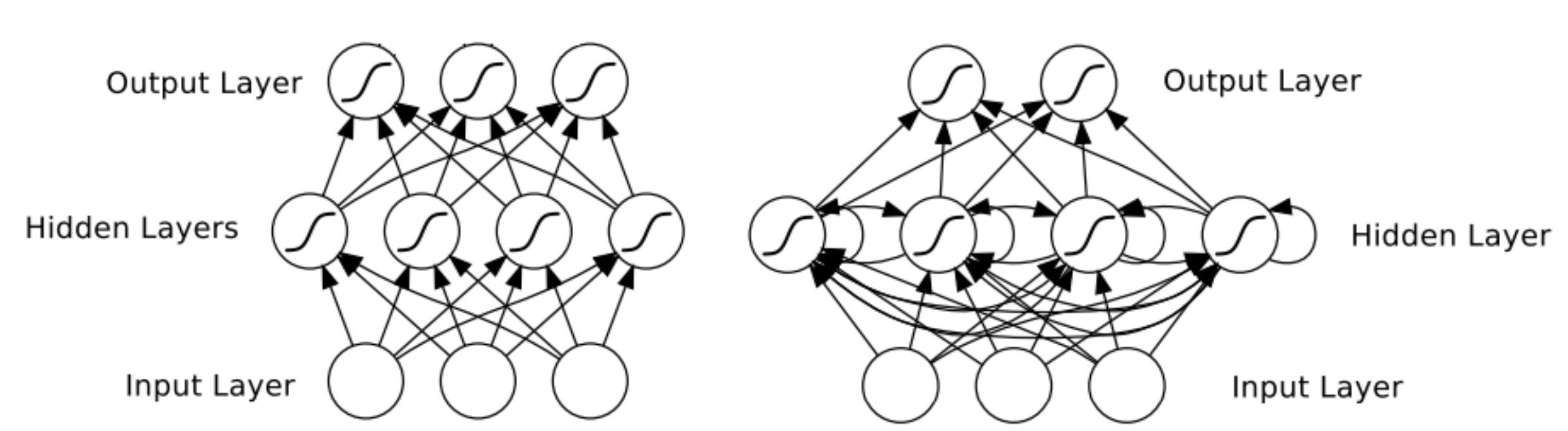} \\
(a) \hspace{4cm} (b) 
\end{tabular}
\end{center}
\caption 
{ \label{figrnn} Illustration of: \textbf{(a)} a multilayer perceptron and \textbf{(b)} a recurrent neural network.} 
\end{figure*}

\begin{figure}
\includegraphics[width=8.8cm,height=5cm]{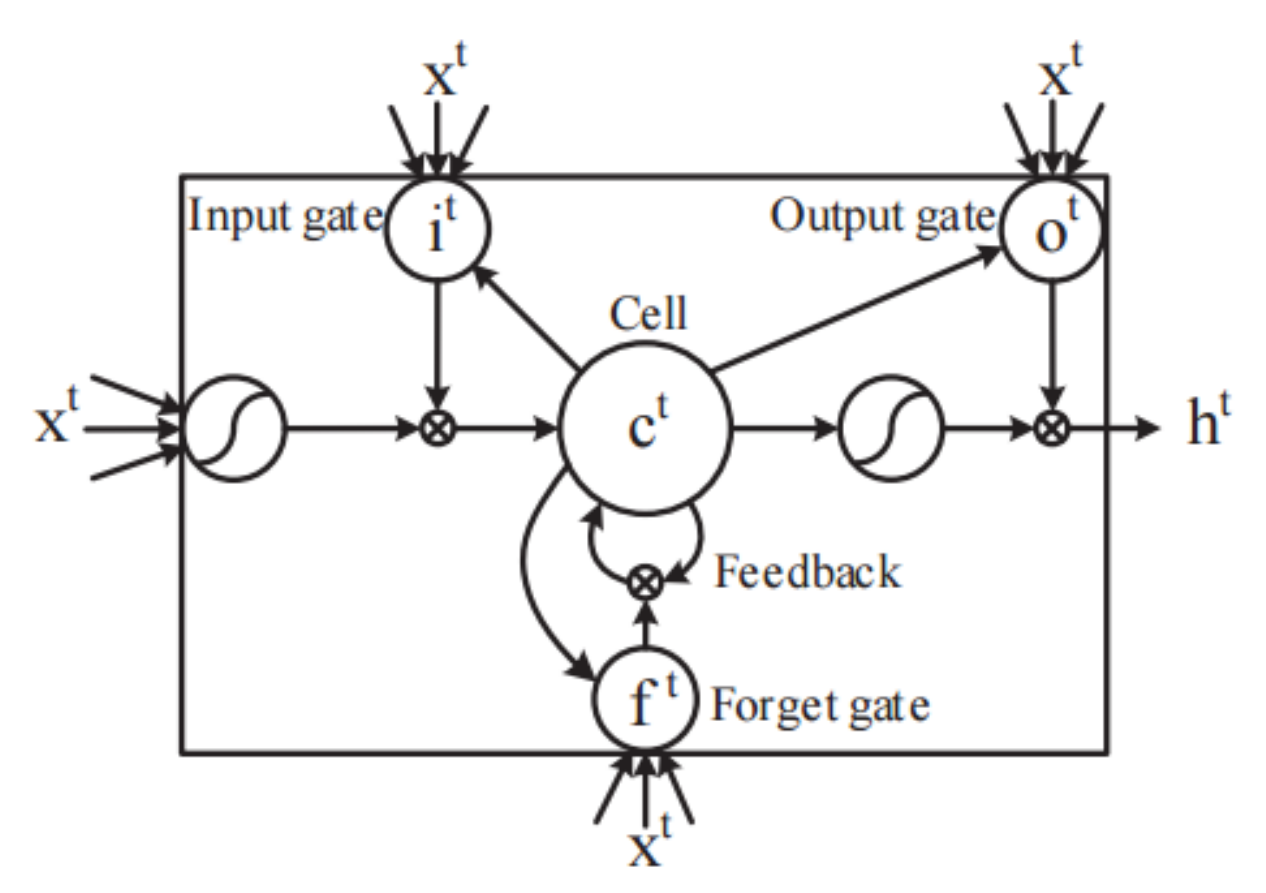} \\
\hspace*{1.1cm} $i^t = \sigma (W_{xi}x^t + W_{hi}h^{t-1} + W_{ci}c^{t-1} + b_i)$ \\
\hspace*{1.1cm} $f^t = \sigma (W_{xf}x^t + W_{hf}h^{t-1} + W_{cf}c^{t-1} + b_f)$ \\
\hspace*{1.1cm} $c^t = f^tc^{t-1} + i^ttanh(W_{xc}x^t + W_{hc}h^{t-1} +b_c)$ \\
\hspace*{1.1cm} $o^t =  \sigma (W_{xo}x^t + W_{ho}h^{t-1} +  W_{co}c^t + b_o  )$ \\
\hspace*{1.1cm} $h^t = o^ttanh(c^t)$
\caption 
{ \label{figlstm} Diagram of an LSTM unit \cite{Graves2008SupervisedSL}. A typical LSTM unit contains an input gate $i^t$, a forget gate $f^t$, an output gate $o^t$, an output state $h^t$ and a memory cell state $c^t$. The information flow is described by the above equations where $\sigma$ is the sigmoid activation; $x^t$ is the input to the network at time $t$; all the matrices $W$ are the connection weights between units. $\odot$ denotes element-wise product; and $u^t$ denotes the modulated input function.  } 
\end{figure}

Recurrent Neural Network (RNN) is a good choice to model the complex dynamics of various actions in video because its architecture allows to store and access the long range contextual information of a temporal sequence. The main difference between an RNN and a multilayer perceptron is the presence of cyclical connections (Figure~\ref{figrnn}). This way, an RNN can learn to map from the entire history of previous inputs to each output \cite{Graves2008SupervisedSL}. However, they are very difficult to train due to the ``\textit{vanishing gradient problem}'' \cite{Bengio1994LearningLD,5264952}. The Long Short-Term Memory (LSTM) approach \cite{hochreiter1997long} has been proposed to solve these problems. Figure~\ref{figlstm} describes the LSTM's structure and its information flow.

\begin{figure}
\begin{center}
\begin{tabular}{c}
\includegraphics[width=8.5cm,height=5cm]{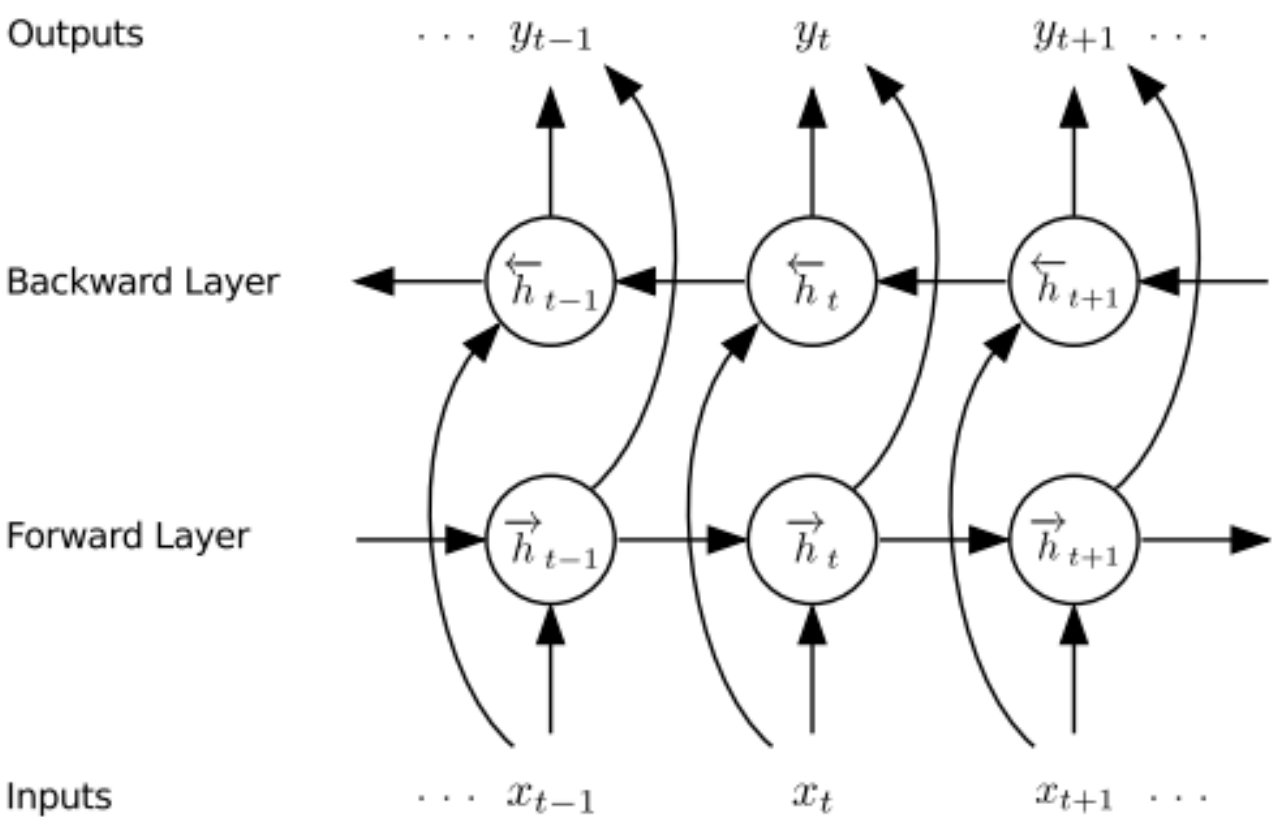}
\end{tabular}
\end{center}
\caption 
{ \label{figbi-lstm} Architecture of a Bidirectional-LSTM. The circular nodes represent LSTM cells.} 
\end{figure}

RNNs not only are able to make use of previous context but also able to exploit future context as well.  Bidirectional RNNs \cite{Schuster1997BidirectionalRN} has been proposed to do this by processing data in both directions with two separate hidden layers. All the information are then sent forwards to the same output layer. By replacing the nonlinear units in the Bidirectional RNNs architecture by LSTM cells, we can obtain Bidirectional-LSTM as shown in Figure~\ref{figbi-lstm}. In subsection~\ref{subsection:4.4}, we will see how to apply Bidirectional-LSTMs to model and recognize human actions in video.

\subsection{Deep belief networks (DBNs)}
\begin{figure*} 
\begin{center}
\begin{tabular}{c}
\includegraphics[width=15cm,height=5cm]{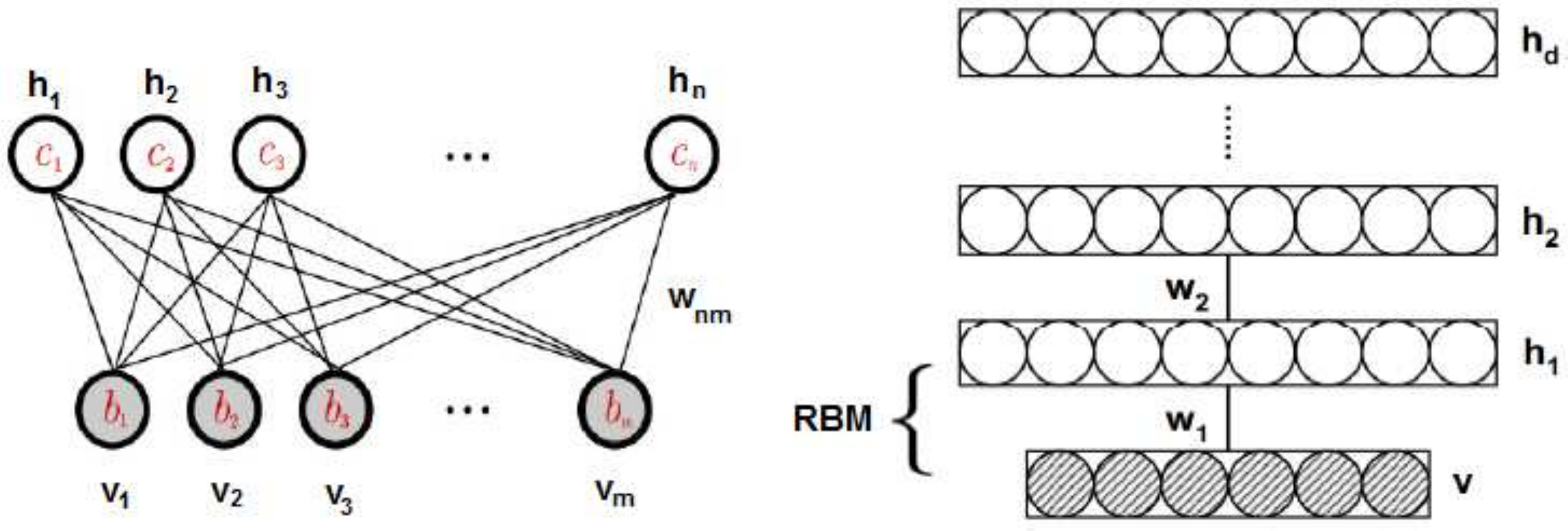}\\
\hspace*{0.3cm} (a) \hspace{7.5cm}(b)
\end{tabular}
\end{center}
\caption 
{ \label{fig10} \textbf{(a)} An example of a RBM with $m$ visible units and $n$ hidden units. \textbf{(b)} The schematic overview of a deep belief networks composed of $d$ RBMs. $W_{1}$,$W_{2}$,...,$W_{h}$ are the weights matrices between the connections.} 
\end{figure*}

DBNs \cite{hinton2006fast} have been used successfully for many recognition tasks such as handwritten digits recognition \cite{hinton2002training}, object recognition \cite{nair20093d}, or modeling human motion \cite{taylor2006modeling}. DBNs are probabilistic generative models that are constructed by stacking several restricted Boltzmann machines (RBMs) \cite{smolensky1986information,hinton1984boltzmann} (Figure~\ref{fig10}b). RBMs are shallow networks containing two layers: one layer of ``\textit{visible}" units that represents the input data and one layer of ``\textit{hidden}" units that learns to represent features. In an RBM architecture, all visible units of the visible layer are connected to all hidden units of the hidden layer, but there are no connections between two units of the same layer (Figure~\ref{fig10}a). The standard type of RBM has binary-valued hidden and visible units, meaning that each unit can only be in one of two states, ``0" or ``1". The probability of setting a unit to ``1" is a sigmoid function of its bias, weights on connections, and the state of other units. More detail, given a binary RBM with $m$ visible units $\mathcal{V} = \{ v_{i} \}, i \in (1,...,m)$ and $n$ hidden units $\mathcal{H} = \{ h_{j} \} , j \in (1,...,n)$, where $v_{i}$ and $h_{j}$ are the binary states of visible unit $i$ and hidden unit $j$ or $(v_{i},h_{j}) \in (0,1)^{m+n}$, the joint probability distribution for visible and hidden units is defined as\cite{hinton2010practical}:
\begin{equation}
P(v_{i},h_{j}) = \dfrac{1}{Z} e^{-E(v_{i},h_{j})}
\end{equation}
where $Z$ is the partition function computed by summing over possible pairs of $(v_{i},h_{j})$:
\begin{equation}
Z = \sum_{v_{i},h_{j}} e^{-E(v_{i},h_{j})}
\end{equation}  and $E(v_{i},h_{j})$ is the energy function given by:
\begin{equation}
\label{eq:3}
E(v_{i},h_{j}) = - \sum_{i = 1}^{m} a_{i}v_{i} - \sum_{j = 1}^{n} b_{j}h_{j} - \sum_{i,j} v_{i}h_{j}w_{i,j}.
\end{equation}
In function \ref{eq:3}, $a_{i}$ and $b_{j}$ are biases, $w_{i,j}$ is the weight between $v_{i}$ and $h_{j}$ units. In a binary RBM model, there are no direct connections between visible units nor between hidden units. So, given the input data $\textbf{v}$ through the visible units, the binary state of each unit $h_{j}$ is 1 with probability:
\begin{equation}
\label{eq:4}
p(h_{j} = 1| \textbf{v}) = \sigma(b_{j} + \sum_{i} v_{i}w_{i,j}).
\end{equation}
Given a hidden vector $\textbf{h}$, we can also reconstruct the states of a visible unit by:
\begin{equation}
p(v_{i} = 1 | \textbf{h}) = \sigma(a_{i} + \sum_{j}h_{j}w_{i,j})
\end{equation}
where $\sigma(x)$ is the sigmoid function with form: $\dfrac{1}{1 + e^{-x}}$.
\\
For estimating the weights $w_{i,j}$ and biases $a_{i}$, $b_{j}$, we use:
\begin{equation}
\dfrac{\partial \log p(\textbf{v})}{\partial w_{i,j}} = \langle v_{i}h_{j}\rangle_{data} - \langle v_{i}h_{j}\rangle_{model}
\end{equation}
\begin{equation}
\dfrac{\partial \log p(\textbf{v})}{\partial a_i} = \langle v_{i}\rangle_{data} - \langle v_{i}\rangle_{model}
\end{equation}
\begin{equation}
\dfrac{\partial \log p(\textbf{v})}{\partial b_{j}} = \langle h_{j}\rangle_{data} - \langle h_{j}\rangle_{model}
\end{equation}\\
The conditional distribution $p(h_{j} | \textbf{v})$ in equation \ref{eq:4} shows that the hidden layer can be constructed by updating the state of units $h_{j}$ when given  a data vector $\textbf{v}$. In practice, since all units in the hidden layer are conditionally independent given the visible layer, the state of each unit can be computed by using block Gibbs sampling \cite{hinton2006fast}. This technique allows to update the state of all the units in parallel. As shown in Figure~\ref{fig10}b, a DBN could be viewed as a stack of several RBMs. Therefore, training a DBN is performed through training each of its RBM. The work of Hinton \textit{et al}. \cite{hinton2006fast} provided an efficient procedure for training DBNs. In this process, the units of the current hidden layer are regarded as visible layer for the next hidden layer and training a DBN starts from the lowest RBM. The procedure is repeated layer-to-layer until the highest RBM is reached and known as the ``\textit{greedy layer-wise training strategy}". Each component (an RBM) of the DBNs acts as a feature extractor on inputs. It extracts ``\textit{low level}" features at the bottom hidden layer, as well as more ``\textit{abstract}" features at the higher hidden layers. To improve the performance of DBNs for classification tasks, the DBN model could be extended by adding a soft-max layer on the top of its architecture.

\subsection{Stacked Denoising Autoencoders (SDAs)}
\label{subsection:3.3}
\begin{figure}
\begin{tabular}{c}
\includegraphics[width=9cm,height=7cm]{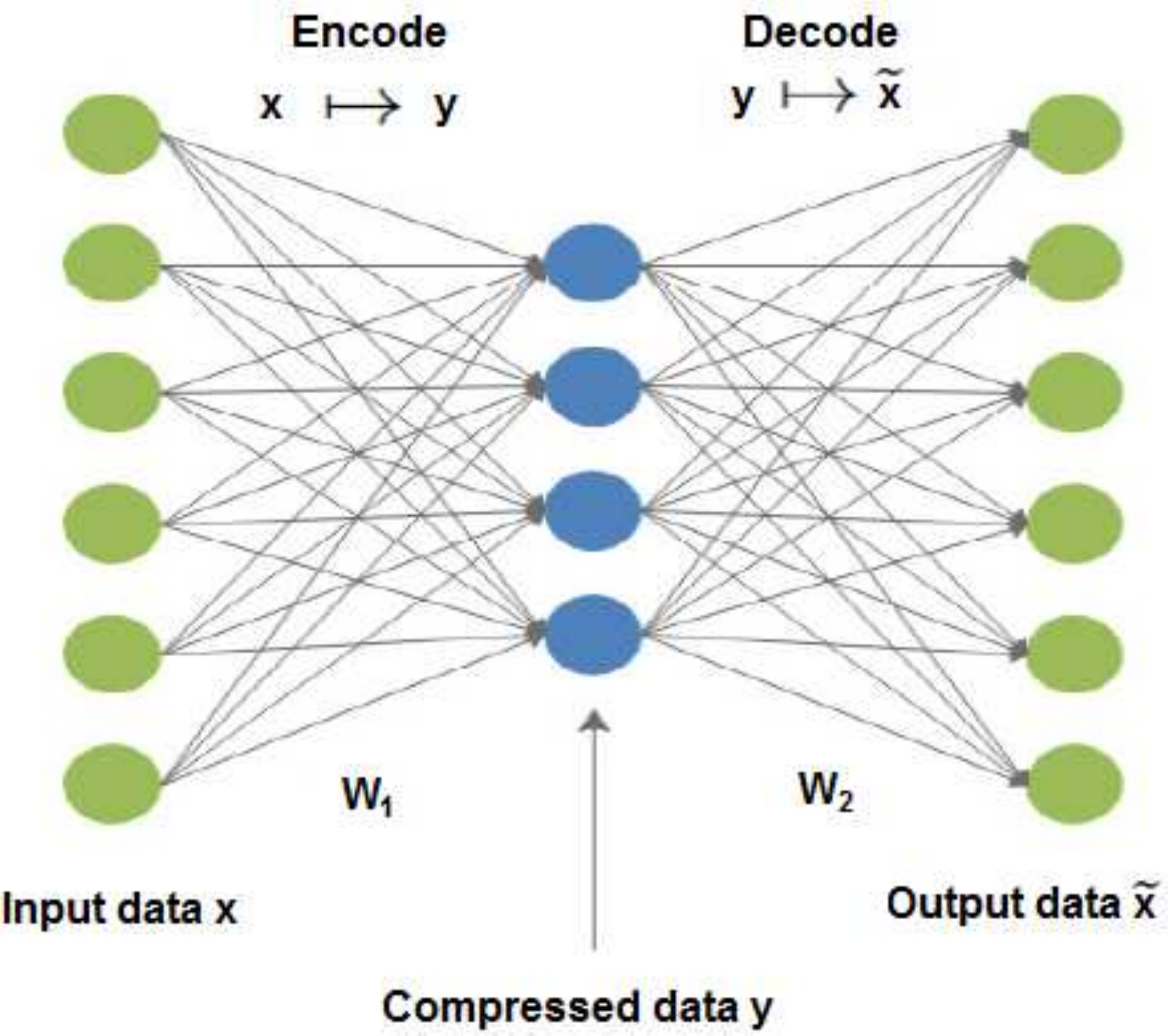}
\end{tabular}
\caption 
{ \label{fig11} The typical structure of an autoencoder.} 
\end{figure}

SDA is another important technique in DL. It is an extension of a classical autoencoder \cite{rumelhart1985learning} and was first introduced in 2008 by Vincent \textit{et al}. \cite{vincent2008extracting}. The idea of an autoencoder is shortly described here: Given a set of data points $\textbf{x} = \{x_{1}, x_{2},...,x_{m}\}$, map $\textbf{x}$ to another set of data points $\textbf{y} = \{y_{1}, y_{2},...,y_{n}\}$ where $n < m$. From the compressed set $\textbf{y}$, we reconstruct a set of $\tilde{\textbf{x}}$, which approximates the original data $\textbf{x}$. The mapping $\textbf{x} \mapsto \textbf{y}$ is called ``\textit{encoding}" and the mapping $\textbf{y} \mapsto \tilde{\textbf{x}}$ is called ``\textit{decoding}". Formally, the processes of encoding and decoding are performed as follows:
\begin{equation}
\textbf{y} = W_{1}x_{i} + b_{1}
\end{equation}
\begin{equation}
\label{eq:10}
\tilde{\textbf{x}} = W_{2}y_{i} + b_{2}.
\end{equation}
where $W_{1} \in \mathbb{R}^{m \times m}, W_{2} \in \mathbb{R}^{n \times n}$. Figure~\ref{fig11} illustrates the network architecture of a typical autoencoder.
To achieve the goal of reconstructing $\tilde{\textbf{x}}$ to approximate the original data $\textbf{x}$, we minimize the difference between $\textbf{x}$ and $\tilde{\textbf{x}}$ by minimizing the function:
\begin{equation}
\label{eq:11}
J (W_{1},b_{1},W_{2},b_{2}) = \sum_{i = 1}^{m} (\tilde{x}_{i} - x_{i})^2.
\end{equation}
From equations \ref{eq:10} and \ref{eq:11}, we have:
\begin{equation}
J (W_{1},b_{1},W_{2},b_{2}) = \sum_{i = 1}^{m} (W_{1}W_{2}x_{i} - 1)x_{i} + b_{1}W_{2} + b_{2})^2.
\end{equation}
SDAs are constructed by stacking several autoencoders together to create a ``\textit{deep}" architecture where the weights are fine-tuned with a back-propagation algorithm \cite{cilimkovic2015neural}. The ''\textit{unsupervised pre-training}'' of each autoencoder is performed in a greedy layer by layer manner. Once the a SDAs is learnt, its output will then be used as the input representation of a supervised learning algorithm for recognition tasks.

\section{Human action recognition approaches based on DL} \label{section:4}

This section reviews current studies of deep learning on human action recognition. We categorized publications based on the proposed taxonomy, including: human action recognition based on CNNs (subsection \ref{subsection:4.1} ); human action recognition based on DBNs (subsection \ref{subsection:4.2} ); human action recognition based on SDAs (subsection \ref{subsection:4.3}); human action recognition based on RNN-LSTMs (subsection \ref{subsection:4.4}), and some other architectures (subsection \ref{subsection:4.5}).

\subsection{Human action recognition based on CNNs}
\label{subsection:4.1}
\begin{figure*}
\begin{center}
\begin{tabular}{c}
\includegraphics[width=15cm,height=6cm]{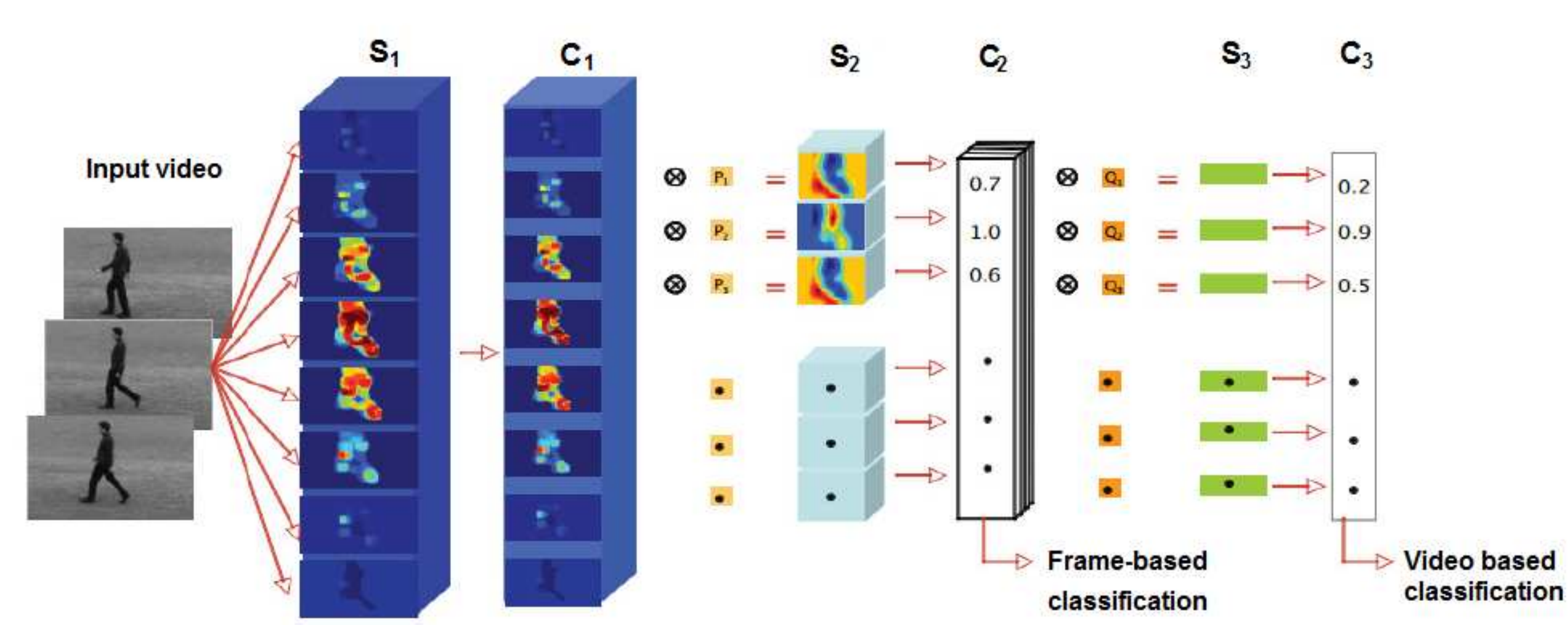}
\end{tabular}
\end{center}
\caption 
{ \label{fig13} The framework for recognizing human action proposed by Jhuang \textit{et al}. \cite{jhuang2007biologically}. Given a gray-value video sequence as input data, the $S_1$  stage locates the object in image frame by using spatio-temporal filters. Each $C_1$ unit is computed by applying a local max over for each $S_1$ unit for down-sampling. From the $C_1$ stage, we perform a template matching operation for identifying intermediate-level features of the model. The $C_2$ stage is constructed by computing the global max over each $S_2$ unit. The high-level features are extracted in $S_3$ through a template matching and the $C_3$ features are computed from $S_3$ using the same way like computing $C_2$. The last stage is a linear multiclass SVM classifier that is able to recognize the actions using the $C_3$ features as input.}
\end{figure*}

Many works on human action recognition and related tasks based on DL models have been proposed and reported in the literature. Among them, one of the most used deep models is CNNs (see subsection \ref{label:cnn}) and its extensions. Researchers have successfully applied CNN-based architectures for many visual tasks such as people detection and tracking \cite{Fan2010HumanTU,sermanet2013pedestrian,wang2015visual}, pose estimation \cite{nowlan1995convolutional,jain2013learning,jain2014modeep,gkioxari2014r,tompson2014real,cheron2015p}, action recognition \cite{giese2003neural,sigala2005learning,jhuang2007biologically,kim2007human,ji20133d,simonyan2014two,wang20143d,wang2015action,tran2015learning,wang2015deep,DOBHAL2015178,liu2015learning,caoaction,mo2016human,singhfirst,pham2018exploiting,huy2018exploiting,pham2020unified,pham2019learning,pham2019spatio,pham2018skeletal,pham2017learning,pham2019deep,pham2019architectures,pham2019skeletal}, event detection and crowded scene understanding \cite{gan2015devnet,shao2015deeply,castro2015predicting,xiong2015recognize}. Early work on applying CNNs was made in 1995 by Nowlan \textit{et al}. \cite{nowlan1995convolutional} for hand tracking and recognizing. In their work, a CNN model is proposed to locate the hand and recognize whether it is close or open with accuracies of 99.7\% and 99.1\% on a dataset of 900 video images from 18 different subjects for each task. However, the complex structured backgrounds of images may have a significant impact on the recognition accuracy. Starting from the work of Fukushima \cite{Fukushima1980}, Giese and Poggio \cite{giese2003neural} proposed a hierarchical feedforward architecture for the recognition of biological movements such as walking, running or various full-body actions. In a related paper, Sigala \textit{et al}. \cite{sigala2005learning} also developed a hierarchical model for detecting a walker based on the use of the neural detectors that are able to extract motion features with different levels of complexity. Jhuang \textit{et al}. \cite{jhuang2007biologically} proposed an extension model from the work of Giese and Poggio \cite{giese2003neural} for the recognition of actions from video sequences (Figure~\ref{fig13}).

In 2007, Kim \textit{et al}. \cite{kim2007human} used a modified CNN model and a weighted fuzzy min-max neural network (WFMM)\cite{kim2006weighted} for human action recognition. In their paper, the CNN generates a set of feature maps from the pretreated data and a WFMM \cite{kim2006weighted} plays the role of a classifier. Normally, the CNNs have been primarily applied on two-dimensional data (2D-CNN) in which these models compute features from the spatial dimensions only. In order to exploit the temporal information of human motion, Ji \textit{et al}. \cite{ji20133d} presented a novel three-dimensional convolutional neural network (3D-CNN) architecture for recognizing human action. This architecture used 3D kernels in the convolution stages to extract motion features from both spatial and temporal dimensions. This improvement can be applied to contiguous frames in video to extract multiple features.

Experiments on TRECVID-2008 \cite{TRECVID-2008} datasets have shown that this model outperforms the frame-based 2D-CNN model and two other methods proposed by Lazebnik \textit{et al}. \cite{lazebnik2006beyond} and Yang \textit{et al}. \cite{yang2009human} which follow the state-of-the-art bag-of-words (BoW) \cite{law2014bag}. Motivated by Ji \textit{et al}. \cite{ji20133d}, Wang \textit{et al}. \cite{wang20143d} has also used 3D-CNN for building a deep architecture for human activity understanding using RGB-D data. In addition, Tran \textit{et al}. \cite{tran2015learning} investigated in detail the 3D-CNN model and showed that it outperforms the 2D-CNN in modeling human motion information on various recognition tasks. Moreover, Tran \textit{et al}. \cite{tran2015learning} found that the best kernel length for 3D-CNN is  $3 \times 3 \times 3$ size. Varol \textit{et al.} \cite{Varol2016LongtermTC} also used 3D-CNN for learning action representation in video but with long-term temporal convolutions at the input layer. This study demonstrated that this solution can significantly improve the performance on state-of-the-art action recognition datasets. A visible disadvantage of 3D-CNN model is the increasing number of parameters of the network. To reduce the complexity of the model, Sun \textit{et al.} \cite{Sun2015HumanAR} proposed a factorized spatio-temporal convolutional network that factorizes the 3D convolution kernels into 2D spatial kernels and followed by 1D temporal kernels.

After finding more efficient ways to train CNNs using GPU computing \cite{1575717} and the success of AlexNet \cite{NIPS2012_4824} in the ILSVRC-2012 competition, much work on human action recognition has been published. Ijjina \textit{et al}. \cite{ijjina2014human} recognize human actions in videos by using the standard action bank \cite{sadanand2012action} as a feature detector and a CNN as a classifier. Gkioxari \textit{et al}. \cite{gkioxari2014r} gave state-of-the-art performance for predicting actions on the PASCAL VOC 2012 detection and action train set \cite{pascalvoc} by using the same CNN architecture as AlexNet \cite{NIPS2012_4824} and extracting region proposals on input image with R-CNN technique \cite{girshick2014rich}. Ch\'eron \textit{et al}. \cite{cheron2015p} designed a new CNN-based pose descriptor for human action recognition from RGB and optical flow information. Two distinct CNNs with an architecture similar to AlexNet \cite{NIPS2012_4824} have been used.

The two-stream convolutional network proposed by Simonyan and Zisserman \cite{simonyan2014two} has shown strong performance for human action recognition in videos. This model is a two-stream architecture including the spatial stream and the temporal stream where each stream is executed by a CNN. The first stream recognizes actions from a single frame, while the second recognizes actions from motion information of multi-frame optical flow. These two streams are then combined for the classification task. The experimental results show that using multi-frame optical flow for training model allows to achieve very good performance with limited training data. This architecture has been seen as the most effective approach of applying DL to action recognition with limited training data. Inspired by the work of Simonyan and Zisserman \textit{et al}. \cite{simonyan2014two}, many different authors have developed two-stream convolutional networks for solving action recognition problems, e.g., Wang \textit{et al.} \cite{wang2016two,wang2016temporal,Wang_UntrimmedNets}, Xiong \textit{et al.} \cite{xiong2016cuhk}. Unlike the two-stream architecture developed by Simonyan and Zisserman \textit{et al}. \cite{simonyan2014two}, Liu \textit{et al}. \cite{liu2015learning} added a module called stCNN (Spatio-Temporal Convolutional Neural Network) to the standard CNN model for exploiting motion and content-dependent features concurrently. Experiments on KTH \cite{schuldt2004recognizing} and UCF-101 \cite{soomro2012ucf101} datasets showed that the recognition accuracy for motion-content combined was better when compared with motion alone. Singh \textit{et al}. \cite{singhfirst} addressed the problem of understanding egocentric activities by using a three-stream CNN architecture. More specifically, the authors proposed a framework for the recognition of wearer’s actions. First, a CNN model called ``\textit{Ego Convnet}" is trained for learning features from egocentric cues including hand mask, head motion, and a saliency map. Then, Ego Convnet is extended by adding two more streams corresponding to spatial and temporal streams as the model proposed by Simonyan and Zisserman \textit{et al}. \cite{simonyan2014two}. Experiments showed that the model with the Ego Convnet stream alone achieved state-of-the-art accuracy on different egocentric videos datasets. In addition, the three-stream architecture.

\begin{figure*}
\begin{center}
\begin{tabular}{c}
\includegraphics[width=15cm,height=5cm]{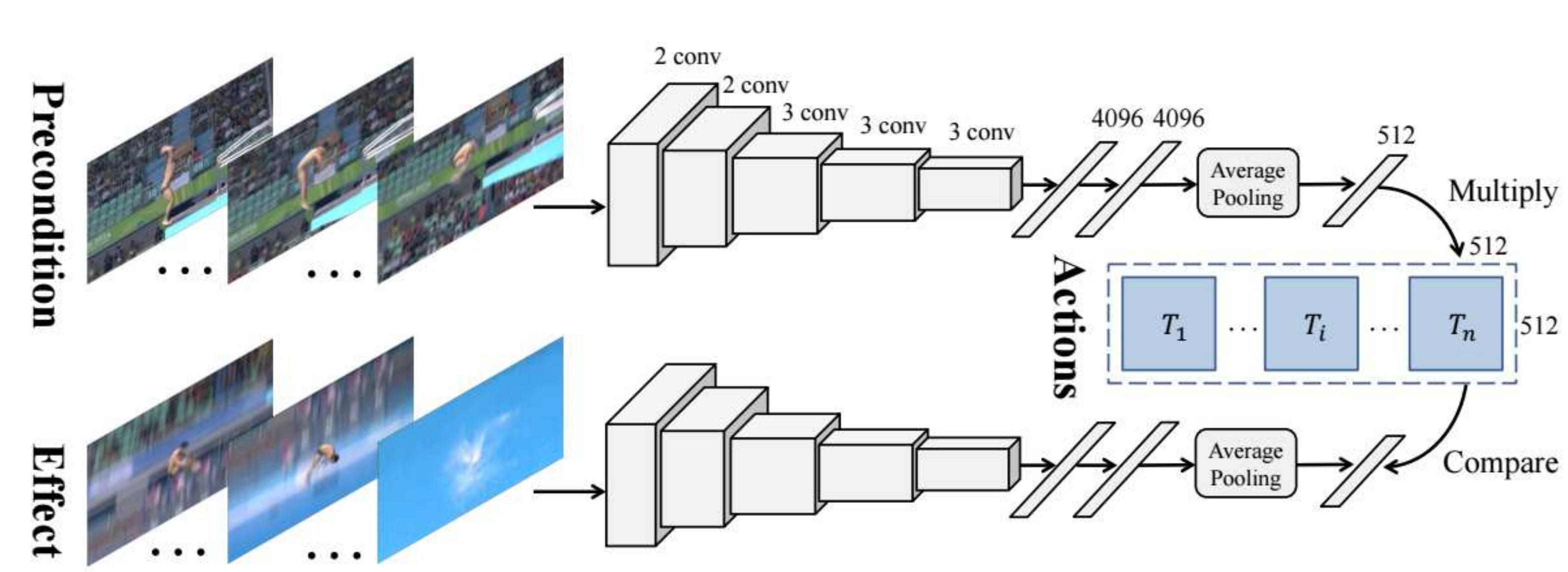}\\
(a) \hspace{3.5cm} (b) 
\end{tabular}
\end{center}
\caption 
{ \label{fig-Siamese-network-architecture}  The Siamese network architecture proposed by Wang \textit{et al.} \cite{Wang2016ActionsT}.}
\end{figure*}

In a recent study, Wang \textit{et al.} \cite{Wang2016ActionsT} divided an input video consisting of $t$ frames $ \mathcal{X} = \{x_1, x_2, ..., x_t\}$ into two sets: the precondition state frames $\mathcal{X}_p = \{x_1, ...,x_{z_p} \}$ and effect state frames $ \mathcal{X}_e = \{x_{z_e}, ...,x_t\}$. The Siamese network architecture has been designed for learning action features. In fact, this is a two-stream CNN models where the first stream is trained on the precondition state frames and the second is trained on the effect state frames as shown in Figure~\ref{fig-Siamese-network-architecture}.

\begin{figure*}
\begin{center}
\begin{tabular}{c}
\includegraphics[width=18cm,height=8cm]{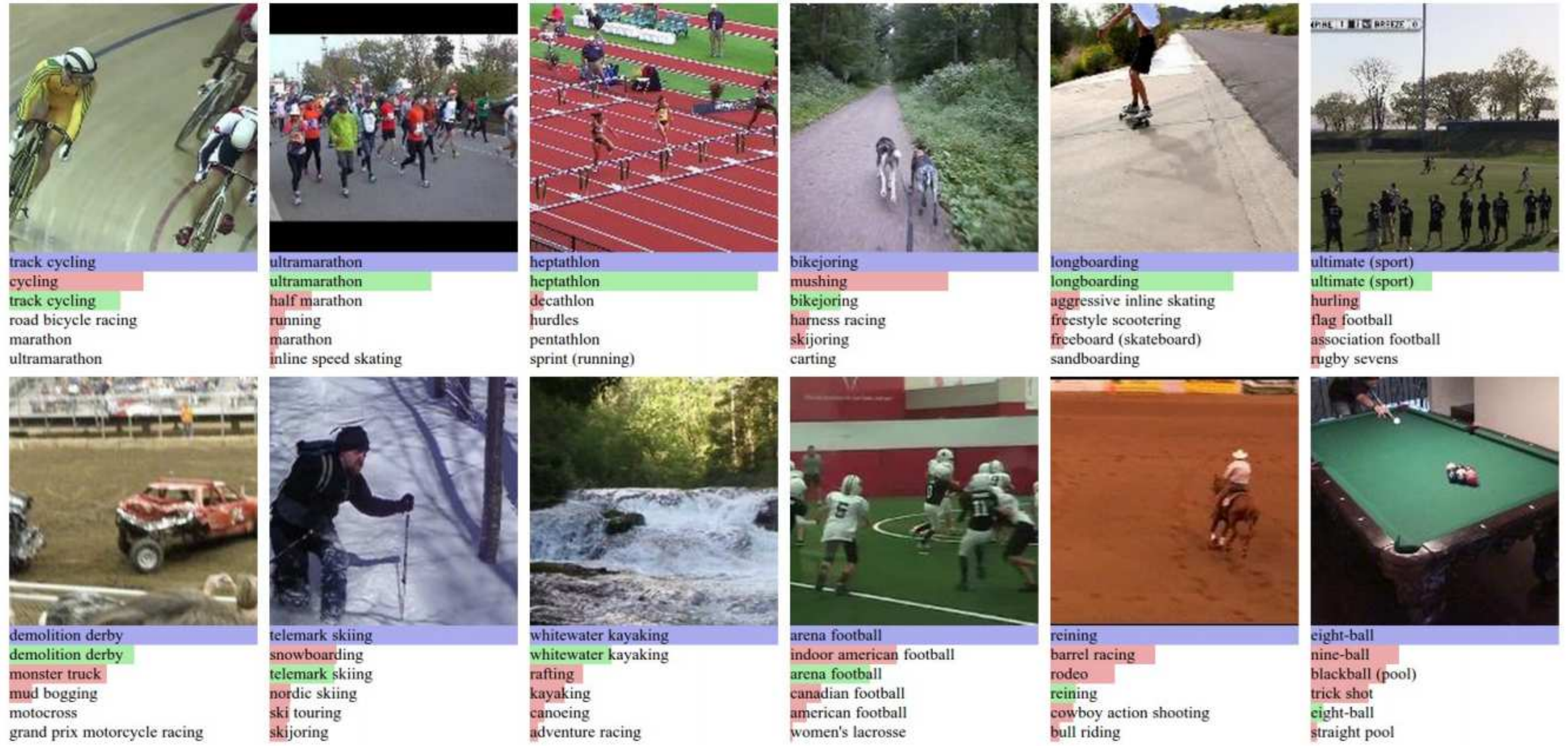}
\end{tabular}
\end{center}
\caption 
{ \label{figmultiresolution} Action prediction on Sports-1M dataset \cite{sport-1m}. The first row indicates ground truth label and the bars below show model predictions. Green and red distinguish correct and incorrect predictions, respectively. \cite{karpathy2014large}.}
\end{figure*}

Advances of 3D sensors such as Microsoft Kinect \cite{zhang2012microsoft} brings up new opportunities in computer vision, even though they tend to be limited to small indoor environments. RGB-D data is able to provide additional information about human motion. Take advantage of depth maps provided by Kinect sensors, Wang \textit{et al.} \cite{Wang:2015:CAR:2733373.2806296} proposed the use of CNNs to learn actions from sequences of depth maps. Given a sequence of depth maps, 3D points are created and three Depth Motion Maps (DMMs) are constructed by projecting the 3D points to the three orthogonal planes. Three CNNs are constructed based on AlexNet architecture \cite{NIPS2012_4824} to extract motion features from each DMM and then classify them into classes. This study is extended in \cite{7358110} and \cite{wang2015deep}. State-of-the-art results have been shown on MSR Action3D Dataset \cite{li2010action}, an extension of the MSR Action3D Dataset, UTKinect-Action Dataset \cite{xia2012view}, and MSR-Daily-Activity3D Dataset \cite{wang2012mining}. Dobhal \textit{et al}. \cite{DOBHAL2015178} also used depth information and a CNN for recognizing human activities. Given a sequence of 2D images, background subtraction is performed. All binary frames are then stacked into a single image called Binary Motion Image (BMI) which contains the flow of the action and is used as the input for the CNN in training and testing phases. The CNN's architecture is same the architecture introduced by LeCun and Bengio  \cite{LeCun1995ConvolutionalNF}. Their approach is extended for extracting BMI from 3D depth maps and achieved competitive performance on Weizmann \cite{ActionsAsSpaceTimeShapes_pami07} and MSR Action3D Dataset \cite{li2010action}. The key ideas behind CNNs such as ``\textit{local connections}" or ``\textit{shared weights}" and the improvements on GPU computing technology have enabled CNNs to train on very large scale datasets. Karpathy \textit{et al}. \cite{karpathy2014large} studied the performance of CNNs by trying to predict and classify on Sports-1M \cite{sport-1m} dataset which consists of more than one million sport videos. Multiresolution CNN architecture with two separate streams of processing has been proposed for reducing training time. The results show that CNNs are capable of learning powerful features and significantly outperform the feature-based baseline. Figure~\ref{figmultiresolution} shows some examples of predictions on Sports-1M dataset \cite{sport-1m}.

In addition to RGB-D information, the acquisition of the skeleton data has become easier with the support of RGB-D sensor. Mo \textit{et al}. \cite{mo2016human} presented a deep model which combines a CNN with a multilayer perceptron \cite{ruck1990feature} for recognizing the human activities based on skeleton data acquired from a Kinect sensor \cite{zhang2012microsoft}. The method achieves a recognition accuracy of 81.8\% on the CAD-60 dataset \cite{CAD_60_database}. Skeleton data has been used by Wang \textit{et al.} \cite{DBLP:journals/corr/WangLLH16}. Firstly, the spatio-temporal information of the joint trajectories is encoded into color images. Then, a CNN based on the AlexNet architecture \cite{NIPS2012_4824} is used to learn the color distribution and to classify actions. The idea of encoding the spatio-temporal information of a skeleton sequence into color texture images and using a standard CNN architeture such as AlexNet \cite{NIPS2012_4824} can also be found in the work of Hou \textit{et al.} \cite{7742919}.

Among the local space-time features, trajectories are one of the best ways to describe motion \cite{wang2011action,wang2013action,beaudry2016efficient}. Wang \textit{et al}. \cite{wang2015action} combined the benefits of improved trajectories \cite{wang2013action} and two-stream CNN architecture from the work of Simonyan and Zisserman \textit{et al}. \cite{simonyan2014two} for designing an effective representation of video feature called ``\textit{Trajectory-Pooled Deepconvolutional Descriptor (TDD)}". The experimental results show that this framework has obtained state-of-the-art performance for recognizing action on the UCF-101 \cite{soomro2012ucf101} and HMDB51 datasets \cite{kuehne2011hmdb}. Inspired by the work of Wang \textit{et al}. \cite{wang2015action}, Cao \textit{et al}. \cite{caoaction} proposed a novel 3D deep convolutional descriptor based on joint positions named ``\textit{Joints-Pooled 3D Deep Convolutional Descriptors (JDD)}". Promising experimental results on sub-JHMDB \cite{jhuang2013towards}, Penn Action \cite{zhang2013actemes}, and Composable Activities \cite{lillo2014discriminative} have shown that using joint-based descriptor with deep model is an effective and robust way for understanding human action.
A new powerful and simple representation of videos for action recognition based on DL, especially CNNs,  called ``Dynamic Image'' has been presented in the work of Bilen \textit{et al.} \cite{Bilen2016DynamicIN}. The idea of this paper is summarizing the video content in a single standard RGB image, then using a pre-trained CNN model such as AlexNet \cite{NIPS2012_4824} on a dataset of dynamic images with fine-tuning technique. The authors also proposed to train CNN from scratch by generating more dynamic images from video segments. Experiments on HMDB-51 and UCF-101 datasets shown the effectiveness of the ``Dynamic Image'' representation.

 \begin{figure*}
\begin{center}
\begin{tabular}{c}
\includegraphics[width=15cm,height=7cm]{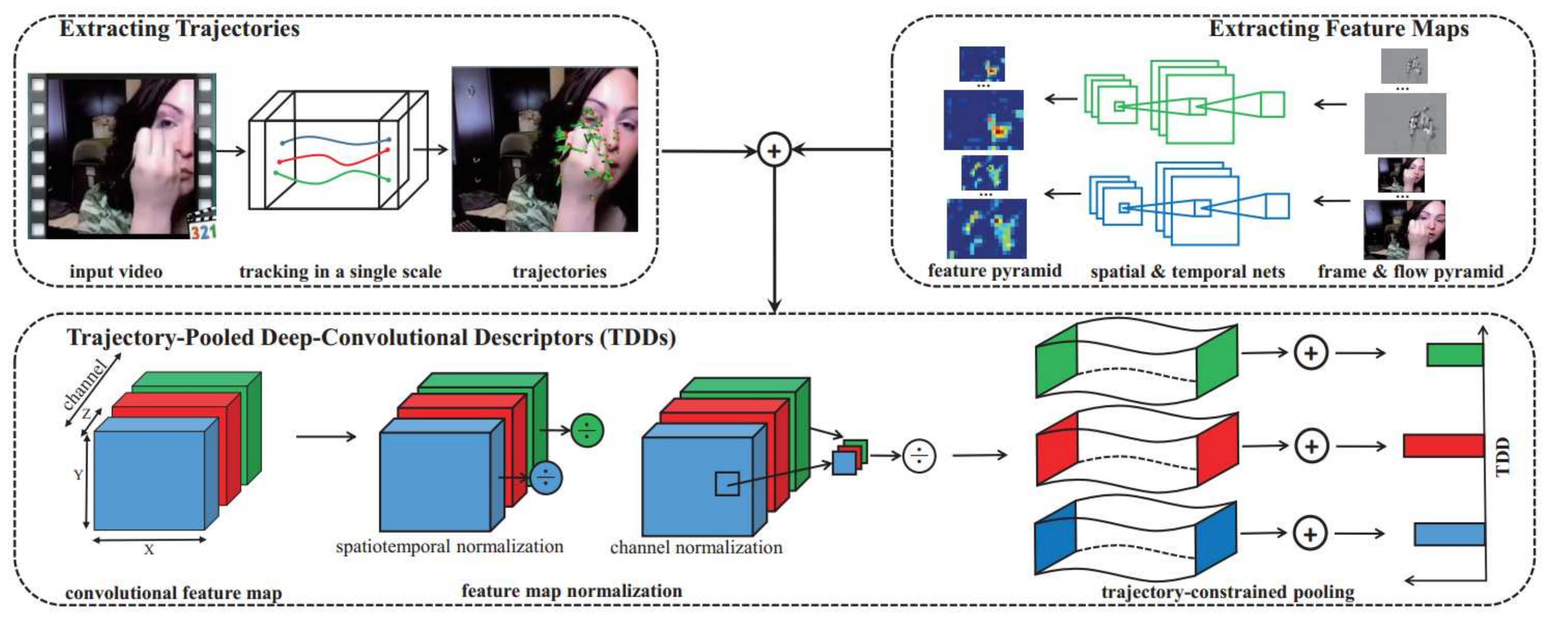}
\end{tabular}
\end{center}
\caption 
{ \label{fig20} The framework for action recognition proposed by Wang \textit{et al}. \cite{wang2015action}. Given an input video, the model extracts trajectories. Multiscale convolutional feature maps are extracted by a CNN at the same time. Trajectory Pooled deep-Convolutional Descriptors (TDDs) are then estimated from a set of improved trajectories and convolutional feature maps.}
\end{figure*}

Very deep convolutional neural networks such as VGGNet \cite{Simonyan2014VeryDC}, GoogLeNet \cite{Szegedy2015GoingDW} have achieved significant success for object recognition and classification tasks. Several authors started to exploit these architectures for action recognition problems. Wang \textit{et al.} \cite{Wang2015TowardsGP} introduced very deep
two-stream CNNs for action recognition based on VGG-16 (VGGNet C with 13 convolutional layers and 3 fully-connected layers) and GoogLeNet \cite{Szegedy2015GoingDW} with 22-layers network. Feichtenhofer \textit{et al.} \cite{Feichtenhofer2016ConvolutionalTN} proposed a CNNs-based novel architecture for spatio-temporal fusion of two stream networks in which the deep CNN model VGG-M-2048 \cite{Chatfield2014ReturnOT} and very deep model VGG-16 \cite{Simonyan2014VeryDC} have been used. The performance comparison between deep (VGG-M-2048) and very deep (VGG-16) models on UCF-101 and HMDB-51 datasets shown that the use of deeper networks improves performance. In addition, GoogLeNet \cite{Szegedy2015GoingDW}  and VGGNet \cite{Simonyan2014VeryDC} have also  been used to design the two-stream CNNs in the work of Wang \textit{et al.}\cite{wang2015cuhk}. Fernando \textit{et al.} \cite{Fernando2016DiscriminativeHR} trained VGG-16 \cite{Simonyan2014VeryDC} on HMDB-51  \cite{kuehne2011hmdb}, UCF-101 \cite{soomro2012ucf101} and Hollywood2 \cite{marszalek2009actions} datasets for obtaining VGG-16 CNN features. The CNN feature vectors are then encoded by a method called  ''hierarchical rank pooling''. This method allows encoding the temporal dynamics of a video sequence for action recognition. A video sequence is encoded at multiple levels in which the output of the each level is a sequence of vectors which captures higher-order dynamics of its previous level. The final representation can be used to learn an SVM classifier for activity recognition as descriptors.
\begin{table}[h!]
  \centering
  \caption{\textbf{Performance comparison of deep model VGG-M-2048 with very
deep model VGG-16 on the UCF-101 \cite{soomro2012ucf101} and HMDB-51 reported by Feichtenhofer \textit{et al.} \cite{Feichtenhofer2016ConvolutionalTN}.}}
  \label{tab6}
  \begin{tabular}{p{0.15\linewidth}p{0.35\linewidth}p{0.35\linewidth}}
  \hline
  \textbf{Dataset} & \hspace{1cm} \textbf{UCF101}  &  \hspace{1cm} \textbf{HMDB51} \\ 
  \hline
  \textbf{Model} & VGG-M-2048 \hspace{0.1cm} VGG-16 & VGG-M-2048 \hspace{0.1cm} VGG-16 \\ 
    \hline
    Spatial & \hspace{0.1cm} 74.22\% \hspace{0.8cm} 82.61\% & \hspace{0.1cm} 36.77\% \hspace{0.8cm} 47.06\% \\
    Temporal & \hspace{0.1cm} 82.34\% \hspace{0.8cm} 86.25\% & \hspace{0.1cm} 51.50\% \hspace{0.8cm} 55.23\% \\
    Spatio-Temporal & \hspace{0.1cm} 85.94\% \hspace{0.8cm} 90.62\% & \hspace{0.1cm} 54.90\% \hspace{0.8cm} 58.17\%\\
	\hline
  \end{tabular}
\end{table}

Very recently, the residual learning framework (ResNets) \cite{he2015deep}, a state-of-the-art CNN and the deepest CNN model at the moment has been exploited for human action recognition by Feichtenhofer \textit{et al.} \cite{feichtenhofer2016spatiotemporal}. In the main ResNet paper \cite{he2015deep}, authors have suggested different architectures of ResNet with 18, 34, 50, 101, 152, and 1202 layers. The underlying network with 50 layer ResNet has been used in the work of Feichtenhofer \textit{et al.} \cite{feichtenhofer2016spatiotemporal} to design a two-stream network. Experiments shown a state-of-the-art performance on UCF-101 \cite{soomro2012ucf101} and HMDB51 \cite{kuehne2011hmdb} datasets.


CNNs are also applied for solving more complex tasks related to human action recognition such as event detection, crowd analysis or behavior prediction. Xu \textit{et al}. \cite{xu2015discriminative} proposed a CNN-based approach for event detection on the large scale video datasets, i.e., TRECVID MEDTest 13 \cite{TRECVID-MED-13} and TRECVID MEDTest 14 \cite{TRECVID-MED-14}. The encoding technique is used for improving the performance and the video representation is compressed for reducing the computation costs. Gan \textit{et al}. \cite{gan2015devnet} presented a CNN-based framework called ``DevNet" for detecting events in videos. Shao \textit{et al}. \cite{shao2015deeply} built a large-scale crowd dataset called WWW Crowd Dataset and designed a CNN model to learn and recognize attributes prediction in crowd video. A similar study can be found in the work of Castro \textit{et al}. \cite{castro2015predicting}. Xiong \textit{et al}. \cite{xiong2015recognize} presented a CNN-based approach which contains two-channels CNN for recognizing complex events from static images. This system is able to detect the objects, predict events, and has given a state-of-the-art result on a very large dataset.

\subsection{Human action recognition based on RNN-LSTMs}
\label{subsection:4.4}
\begin{figure}
\begin{center}
\begin{tabular}{c}
\includegraphics[width=8.5cm,height=5.5cm]{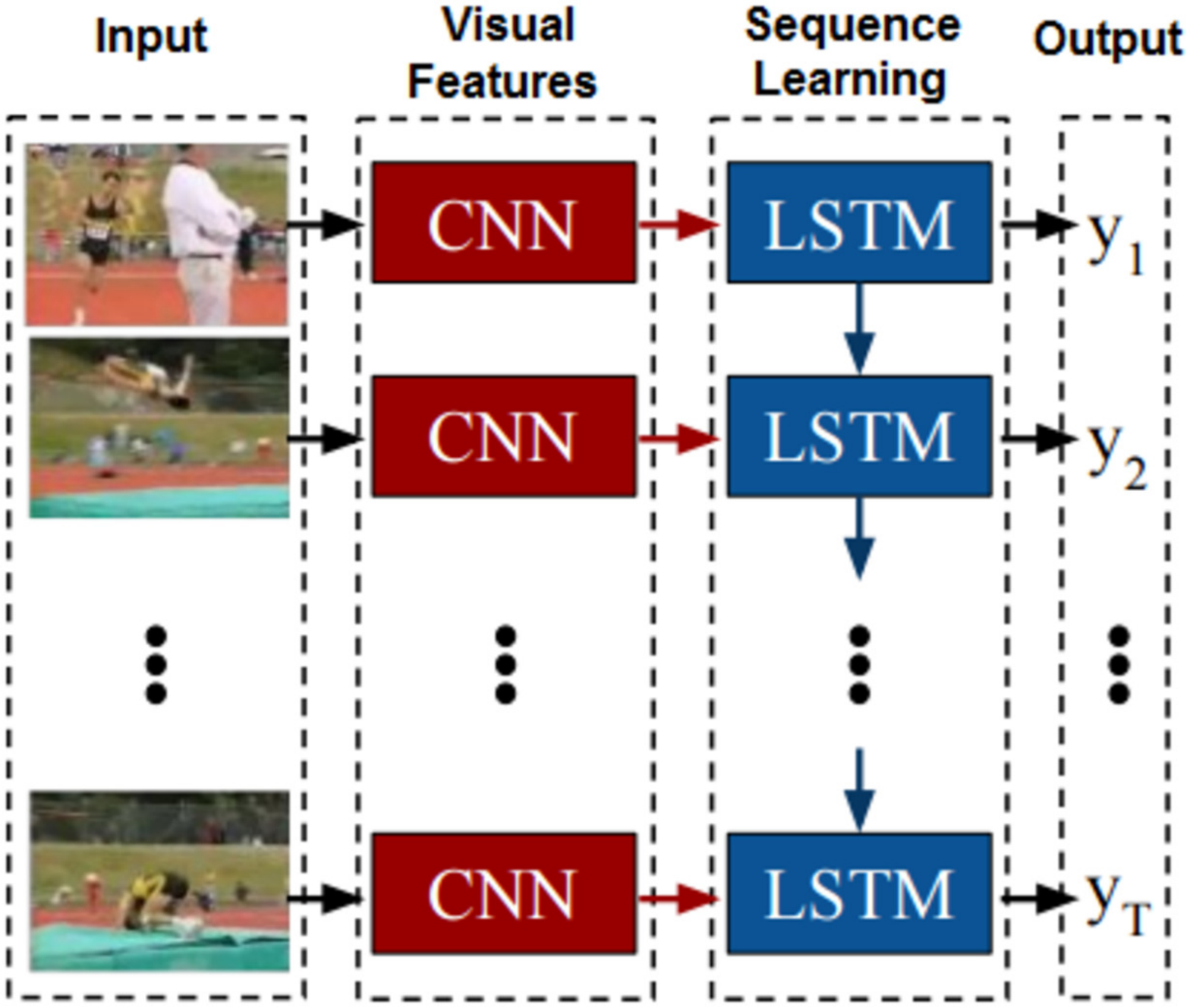}\\ 
\end{tabular}
\end{center}
\caption 
{ \label{lstm-example} Deep learning framework combining CNN and RNN-LSTM for action recognition proposed by Donahue \textit{et al.} \cite{Donahue2015LongtermRC} }
\end{figure}

As pointed out in subsection~\ref{subsection:3.4}, the main advantage of RNN-LSTMs is the capacity to model the long-term contextual information of temporal sequences. This advantage puts RNN-LSTM at one of the best sequence learners for time-series data including visual information of human action. Grushin \textit{et al.} \cite{Grushin2013RobustHA} has demonstrated the robustness of the LSTM network’s performance on the human action recognition task with the hand-crafted feature HOF \cite{laptev2008learning}. As discussed in subsection~\ref{subsection:4.1}, CNNs has been shown its effectiveness in learning features from raw data. Therefore, the works of Baccouche \textit{et al.} \cite{Baccouche2011SequentialDL}, Ng \textit{et al.} \cite{Ng2015BeyondSS}, Donahue \textit{et al.} \cite{Donahue2015LongtermRC}, Giel \textit{et al.}\cite{Giel2015RecurrentNN}, Sharma \textit{et al.} \cite{Sharma2015ActionRU}, Ibrahim \textit{et al.} \cite{Ibrahim2016AHD}, Singh \textit{et al.} \cite{Singh2016AMB}, Li \textit{et al.} \cite{Li2016ActionRB}, Wu \textit{et al.} \cite{Wu2016ActionRW}, Wang \textit{et al.}  \cite{Wang2016HierarchicalAN}, Chen \textit{et al.} \cite{7868164} tackle the question of understanding human actions by combining a CNN and an RNN-LSTM network. The general idea of these papers is to use the standard CNN models such as AlexNet \cite{NIPS2012_4824}, VGGNet \cite{Simonyan2014VeryDC}, or GoogLeNet\cite{Szegedy2015GoingDW} for extracting motion features from input video. Then, RNN-LSTM network is connected to the output of the CNN to classify sequences using learned features. Figure~\ref{lstm-example} shows an example of using CNN and RNN-LSTM for human action recognition from the work of Donahue \textit{et al.} \cite{Donahue2015LongtermRC} . 
While all the work above just uses RNN-LSTMs as a sequence classification, several studies have proposed the use of RNN-LSTMs as an end-to-end learning framework for skeleton based action recognition. E.g., the work of Du \textit{et al.}\cite{du2015hierarchical}, Song \textit{et al.}\cite{Song2016AnES}, Zhu \textit{et al.} \cite{Zhu2016CooccurrenceFL}, Li \textit{et al.} \cite{Li2016OnlineHA}, Liu \textit{et al.}  \cite{Liu2016SpatioTemporalLW}. RNN-LSTMs learn directly motion features and classify them into classes from 3D human-skeleton sequences provided by depth sensors. Experiments on the state-of-the-art datasets demonstrate the effectiveness of these methods.  In another study of Mahasseni \textit{et al.} \cite{Mahasseni2016RegularizingLS} used a parallel architecture to recognize actions with multi-source data. A RNN-LSTM is trained in unsupervised manner on 3D human-skeleton sequences. In the same time, another RNN-LSTM with a CNN is trained on 2D videos. The outputs are then compared to improve the ability of the system.

\begin{figure}
\begin{center}
\begin{tabular}{c}
\includegraphics[width=8.6cm,height=6cm]{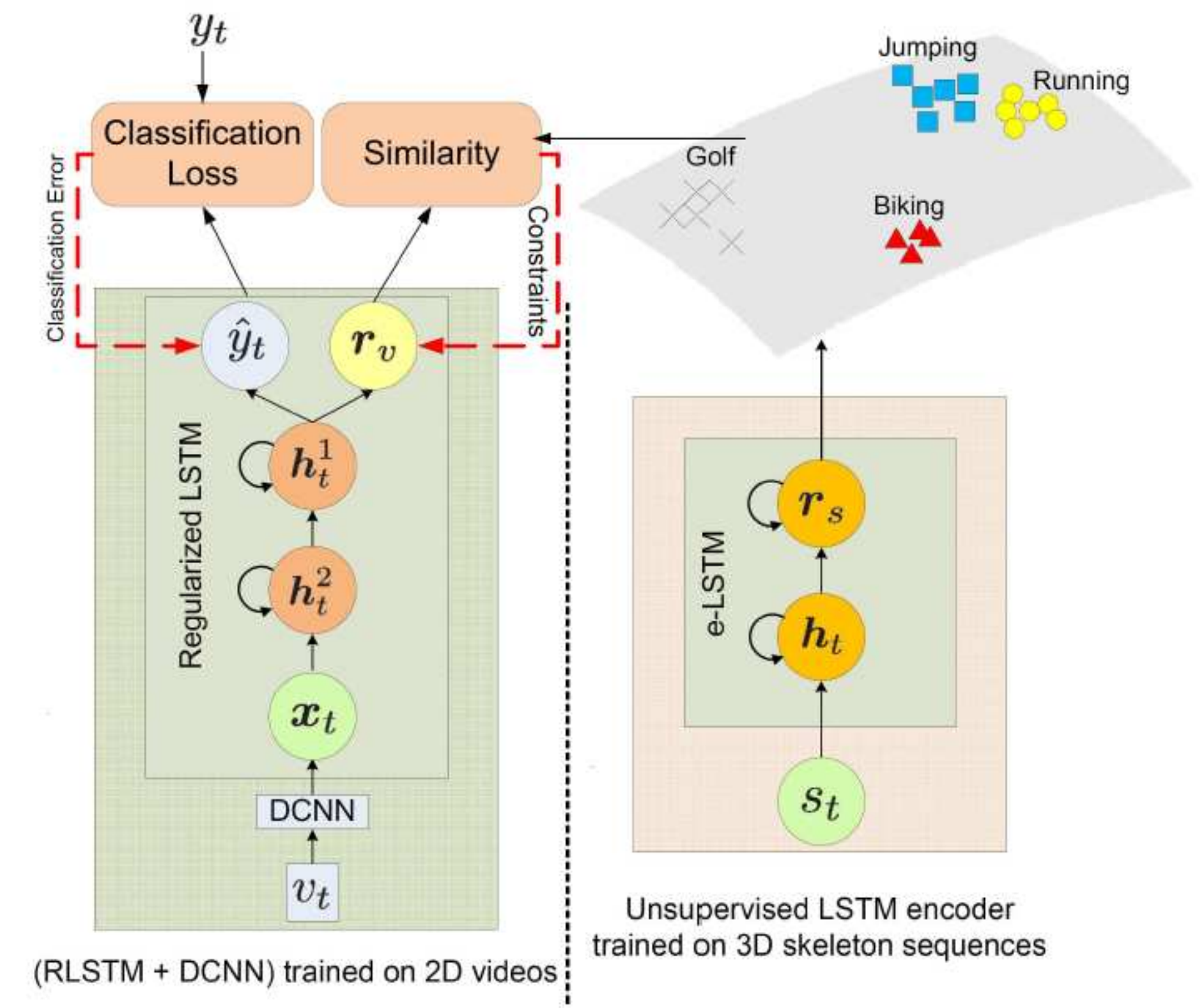}\\ 
\end{tabular}
\end{center}
\caption 
{ \label{cnn-rnn-lstm}  The parallel deep learning architecture with RNN-LSTM proposed by Mahasseni \textit{et al.} \cite{Mahasseni2016RegularizingLS}.}
\end{figure}

\subsection{Human action recognition based on DBNs}
\label{subsection:4.2}

DBNs have become popular DL models after the key paper by Hinton \textit{et al}. \cite{hinton2006fast} presented in 2006. A comparative evaluation by Tang \cite{tangcomparative} showed that DBNs seem ideal for semi-supervised learning, in which we do not need much labeled data. Early work on DBNs was successfully applied for handwritten digits recognition \cite{hinton2006fast} and object recognition \cite{nair20093d,lee2009convolutional}. In 2007, Taylor \textit{et al}. \cite{taylor2006modeling} extended the RBM model by connecting two more visible layers to the hidden layer for modeling human motion. The new model, called the conditional RBM (cRBM) allows to find a single set of parameters that simultaneously capture several different kinds of motion after training on skeleton data. Then, the authors successfully constructed a DBN from cRBMs. Experiments on two motion datasets have demonstrated that this model is able to effectively learn different kinds of motion, as well as the transitions between these kinds.

\begin{figure*}
\begin{center}
\begin{tabular}{c}
\includegraphics[width=15cm,height=5cm]{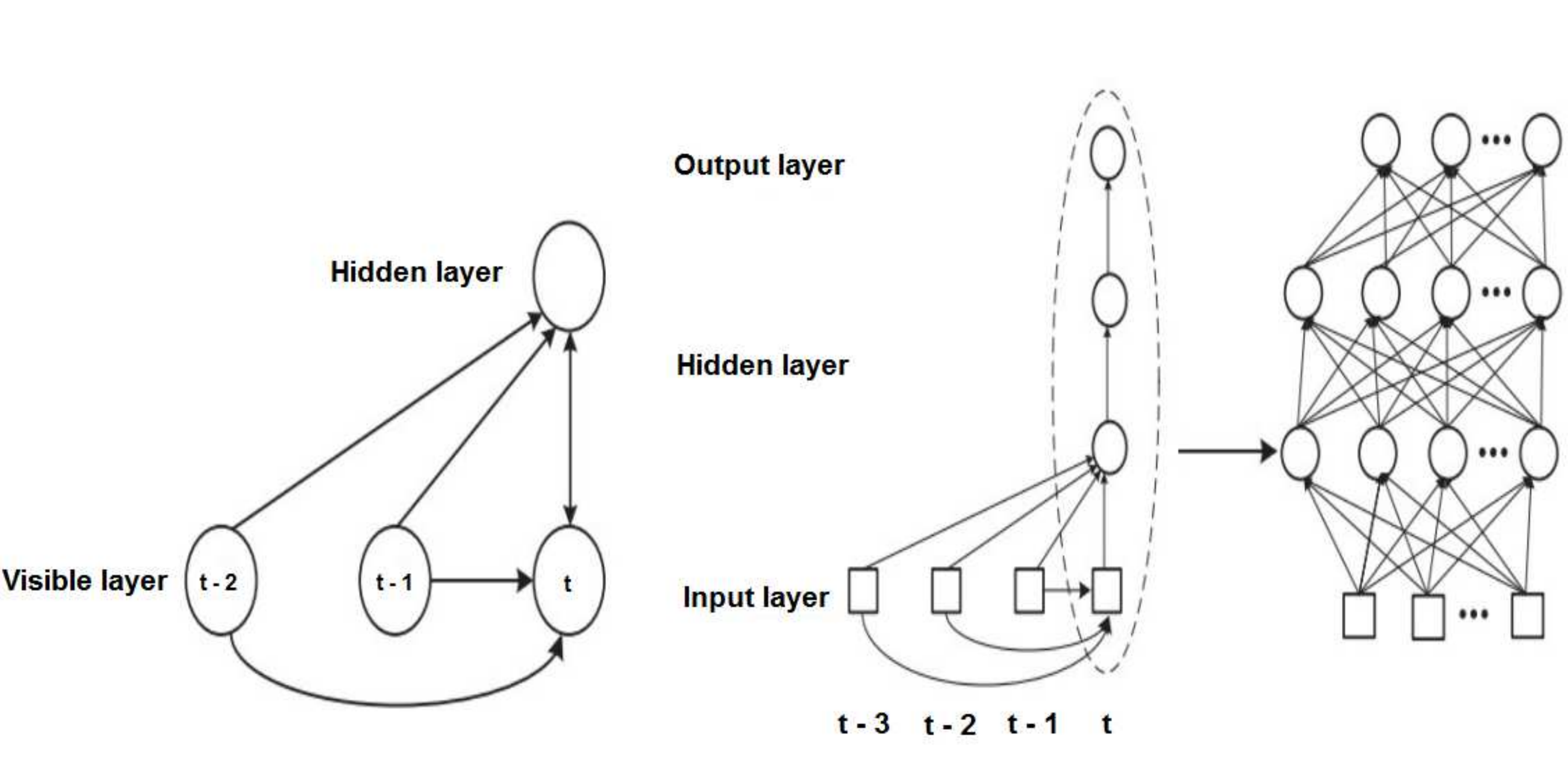}\\
(a) \hspace{7cm} (b) 
\end{tabular}
\end{center}
\caption 
{ \label{fig22} \textbf{(a)} The \textbf{cRBM} model proposed by Taylor et al. \cite{taylor2006modeling}. \textbf{(b)} A modified DBN model designed by Zhang et al. \cite{zhang2014real}.}
\end{figure*}

\begin{figure}
\begin{center}
\begin{tabular}{c}
\includegraphics[width=8.8cm,height=8cm]{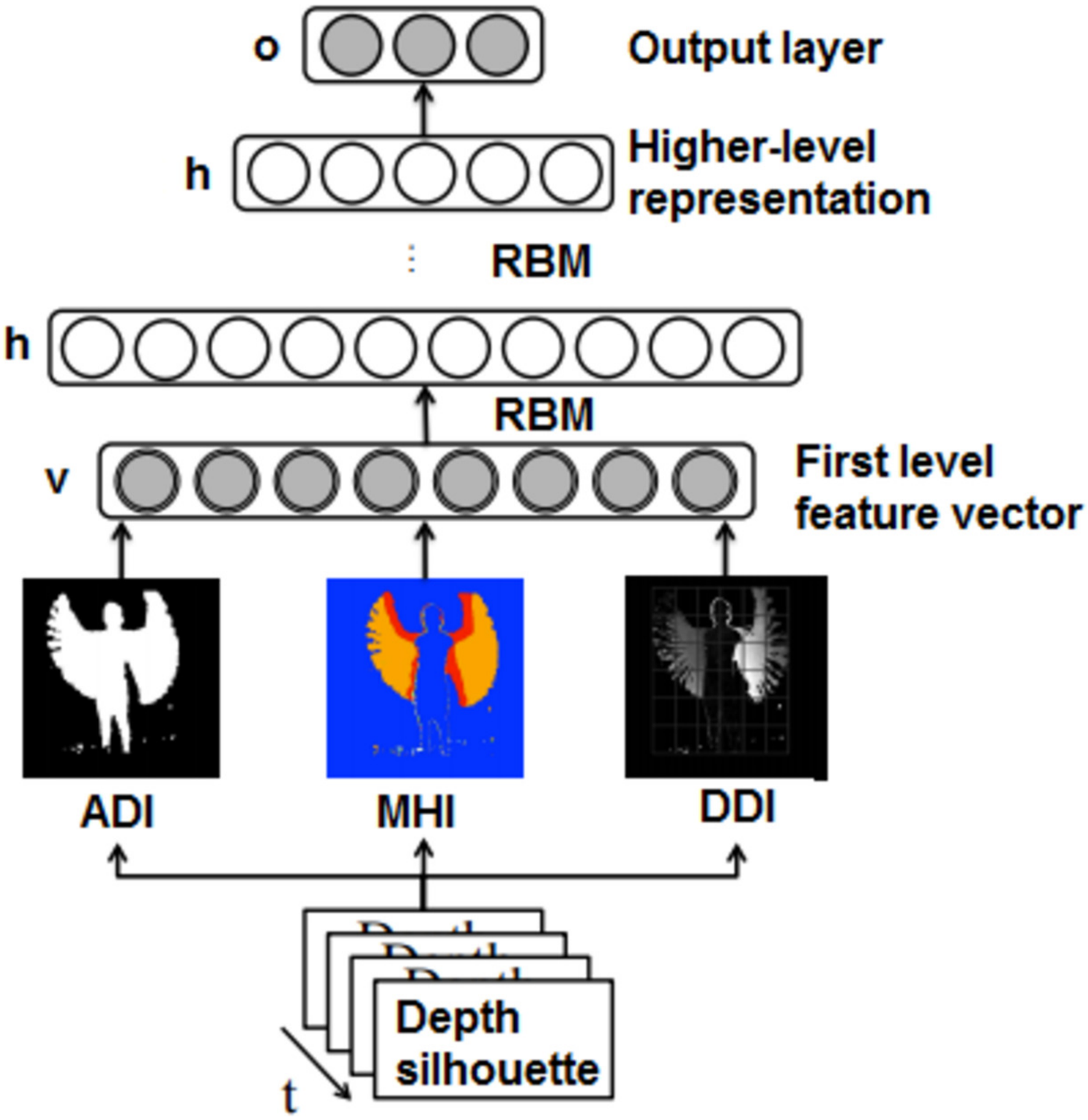}\\ 
\end{tabular}
\end{center}
\caption 
{ \label{fig23} An overview of the DBN architecture for human action recognition proposed by Foggia et al. \cite{foggia2014exploiting}. Three derived images (ADI, MHI,DDI) are computed from depth images and feed into the first level of the network. A more abstract representation is obtained at higher level. Finally, the classification is done using a feed-forward neural network.}
\end{figure}

In another research, Zhang \textit{et al}. \cite{zhang2014real} used a modified DBN model for recognizing human actions in real-time from skeleton data. To achieve this goal, the authors used cRBMs as proposed by Taylor \textit{et al}. \cite{taylor2006modeling} to create the new DBN architecture with two hidden layers as shown in Figure~\ref{fig22}. The proposed model is trained and tested by using the skeletal representation of MSR Action3D \cite{li2010action} and MIT datasets \cite{hsu2005style}. Results show that the recognition accuracy depends on the number of frames. For example, on the MIT datasets \cite{hsu2005style}, the accuracy when using one frame is 98.34\%. Meanwhile, when the number of frames is more than 30, accuracy can reach 100\%. 
Foggia \textit{et al}. \cite{foggia2014exploiting} proposed a DBN-based method for recognizing human actions with depth images. A DBN model is constructed as shown in Figure~\ref{fig23}. Three types of well-known feature including the Average Depth Image (ADI), the Motion History Image (MHI), and the Depth Difference Image (DDI) are computed and encoded as low-level data representation in the first layer. The high level representation is then extracted  by  the proposed model for recognition task.  The achieved results on MIVIA \cite{foggia2013recognizing} and MHAD \cite{ofli2013berkeley} datasets are very promising. 
Ali and Wang \cite{ali2014learning} presented a framework based on DBN to recognize and identify human actions. To speed up learning time, the Fast Fourier Transform (FFT) \cite{heckbert1995fourier} technique is used for converting images to the frequency domain. The model is first pre-trained with KTH dataset \cite{schuldt2004recognizing} and then is used for predicting actions. Experiments showed that the proposed model is better than all published approaches in the literature. More details about this comparison are shown in Table~\ref{tab6}.   
\begin{table}[h!]
  \centering
  \caption{\textbf{Average recognition accuracy of human action on KTH \cite{schuldt2004recognizing} dataset}}
  \label{tab6}
  \begin{tabular}{p{0.8\linewidth}p{0.1\linewidth}}
  \hline
  \textbf{ Approach} & \textbf{Accuracy}  \\ 
    \hline
    DBN by Ali et Wang \cite{ali2014learning} & \textbf{94.3\%} \\
    ISA + Norm-thresholding by Le et al. \cite{Le:2011:LHI:2191740.2192108}  & 93.9\% \\
     Harris3D \cite{laptev2005space} + HOF \cite{laptev2008learning} by Wang \textit{et al.}  \cite{wang2009evaluation} & 92.1\% \\
    Harris3D \cite{laptev2005space} + HOG/HOF \cite{laptev2008learning} by     Wang\textit{ et al.}  \cite{wang2009evaluation}   & 91.8\%   \\
    HMAX \cite{jhuang2007biologically} 			& 91.7\% \\
    3D CNN \cite{ji20133d} 	& 90.2\% \\
    Cuboids \cite{dollar2005behavior} + ISA \cite{laptev2005space} 	& 90.0\% \\
    GRBM \cite{taylor2010convolutional} 	& 90.0\% \\
    Dense + HOF  \cite{laptev2008learning} by Wang\textit{ et al.} \cite{wang2009evaluation}	& 88.0\% \\
    pLSA \cite{niebles2008unsupervised} 	& 83.3\% \\
	Volumetric \cite{ke2005efficient} & 62.7\% \\
   \hline \\
	{\footnotesize Here, accuracy (ACC) is computed as: ACC = $\dfrac{TP+TN}{N}$.}
  \end{tabular}
\end{table}
We can also find in the literature some other human action recognition applications based on DBNs. For example, Nam \textit{et al}. \cite{nam2015single} employed a DBN for developing a real-time human activity recognition using 3D joint positions from RGB-D sensor. The achieved results from these studies confirmed that DBN-based approaches are a good choice for many human action recognition problems.

\subsection{Human action recognition based on SDAs}
\label{subsection:4.3}

As pointed out in subsection \ref{subsection:3.3} SDAs can be trained to reconstruct the input from a corrupted version of it. The first successful application based on the encoder-decoder model is presented in 2007 by Huang \textit{et al}. \cite{huang2007unsupervised} for object recognition tasks. A few years later, based on the principle of the model of Huang \textit{et al}. \cite{huang2007unsupervised}, Baccouche \textit{et al}. \cite{baccouche2012spatio} proposed a solution for learning of sparse spatio-temporal features based on autoencoder scheme. Experiments on KTH \cite{schuldt2004recognizing} and  GEMEP-FERA datasets \cite{valstar2011first} showed the best results when compared to methods using hand-crafted features. Some other autoencoder-based approaches have also been proposed in the works of Wu \textit{et al}. \cite{wu2014adaptive}, Xie \textit{et al}. \cite{xie2014pyramidal}, Hasan \textit{et al}. \cite{hasan2014continuous}, and Budiman \textit{et al}. \cite{budiman2014stacked}. For instance, Wu \textit{et al}. \cite{wu2014adaptive} constructed a 3-layer SDA architecture for human action recognition using skeleton information captured by Kinect \cite{zhang2012microsoft} sensor. Budiman \textit{et al}. \cite{budiman2014stacked} have also performed a similar study when using a SDA model to learn skeleton feature for human body pose classification. To recognize human action, Xie \textit{et al}. \cite{xie2014pyramidal} used a SDA architecture with 3-hidden layers to learn contour features from a single depth frame. Hasan \textit{et al}. \cite{hasan2014continuous} presented an autoencoder-based framework for learning human activity models continuously from streaming videos. This method is executed through two phases: ``\textit{initial learning}" phase and ``\textit{incremental learning}" phase. Given a streaming video with a few labeled activities, the first phase will extract space-time interest points (STIP) \cite{laptev2005space} of the motion then encode these feature vectors by a sparse autoencoder. A softmax function is used as  a classification model that provides action label. To recognize human activities in unlabeled frames, the incremental learning phase uses the sparse autoencoder and the parameters of activity classification model in initial learning phase, but in an unsupervised manner. In this phase, the active learning technique \cite{settles2012active} has also been used to reduce the amount of manual labeling of classes. 

The long training time is a disadvantage of SDAs when working with large-scale datasets. To overcome this limitation, Chen \textit{et al}. \cite{chen2012marginalized} proposed a novel variant of SDAs named ``mSDA". Experiments on the same dataset showed that mSDA matched the performance of SDA but reducing the training time down to 450 times. Taking advantage of the mSDA, Gu \textit{et al}. \cite{gu2015marginalised} trained an mSDA network for multi-view action recognition. An mSDA is trained over all the camera views and the trained network is then used to generate features for each camera view respectively. These obtained features from all the camera views are then combined to create a single integrated representation, which can then be used as the input of a classifier. The evaluation on three benchmark multi-view action datasets provided that this model achieved the state-of-the-art recognition performance.

\subsection{Other deep architectures for human action recognition }
\label{subsection:4.5}

\begin{figure*}
\begin{center}
\begin{tabular}{c}
\includegraphics[width=15cm,height=6cm]{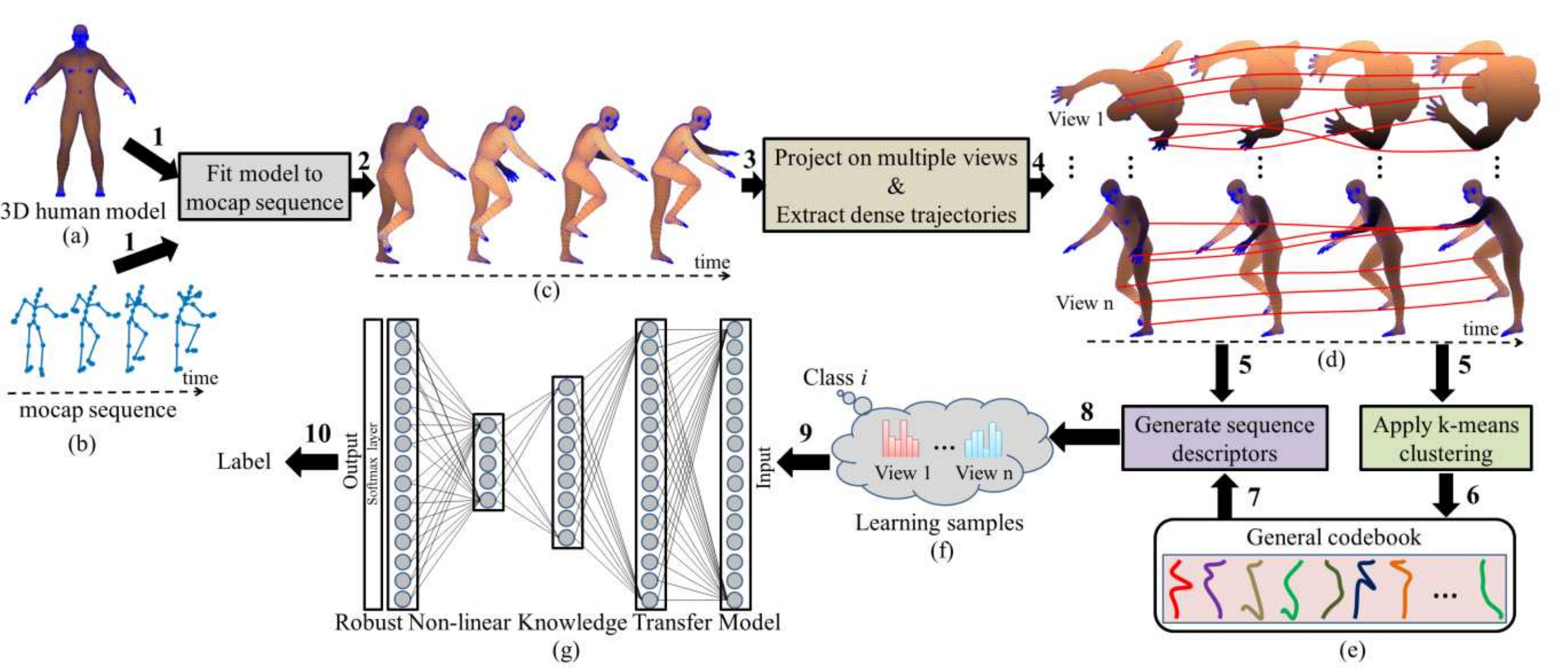}\\ 
\end{tabular}
\end{center}
\caption 
{ \label{fig25} Architeture of R-NKTM and its learning process \cite{rahmani2016learning}. Firstly, 3D human models are fitted to real motion capture data for generating realistic 3D videos. These 3D videos are then projected on 2D planes for calculating dense trajectories. A general codebook is learned from trajectories which is then used as the input of R-NKTM. By this way, the R-NKTM can learn features of human action videos and use it for testing process.}
\end{figure*}

Some other deep architectures have also been used for human action recognition and related recognition tasks such as group activity analysis, or prediction of physical interactions. Sparse coding \cite{olshausen1996emergence,lee2006efficient,yu2011learning} is also another potential deep model for recognizing human action. The success of the sparse representation in various fields including pattern recognition \cite{raina2007self,yang2010supervised} or image classification \cite{yang2009linear} have shown that it could flexibly adapt to diverse low level natural signals. The sparse representations of the signals are then used as image features which are sent directly into the classifiers. Therefore, many authors \cite{zhu2010sparse,lu2011latent,lu2013latent,guha2012learning,alfaro2016action} have exploited the advantages of sparse coding for solving human action recognition problems. Recently, some novel deep architectures for recognizing human action have been published in the literature \cite{ullah2015strict,ullah2016spatiotemporal,rahmani2016learning}. For instance, Ullah and Petrosino \cite{ullah2015strict} employed a CNN and a pyramidal neural network (PyraNet) \cite{phung2007pyramidal} to recognize human action. A strict 3D pyramidal neural network (3DPyraNet) was constructed which allows to learn spatio-temporal features of human motion.  These works continued to be expanded by the same authors \cite{ullah2016spatiotemporal} and achieved competitive results on some action datasets. Rahmani \textit{et al}. \cite{rahmani2016learning} presented the ``\textit{Robust Non-Linear Knowledge Transfer Model}" (R-NKTM), a deep fully-connected neural network which is capable of understanding human action from cross-view by learning features from dense trajectories of synthetic 3D human models and real motion capture data. Figure~\ref{fig25} illustrates the procedure to train this network. 
Experiments on cross-view human action datasets including IXMAS \cite{weinland06}, UWA3DII \cite{rahmani2016histogram}, N-UCLA Multiview Action3D \cite{wang2014cross}, and UCF Sports \cite{4587727} have shown that this method outperforms existing state-of-the-art.

The paper published by Le \textit{et al}. \cite{Le:2011:LHI:2191740.2192108} reports that we can combine the different network models to build a single deep architecture for improving its performance. Based on two key ideas, ``\textit{convolution}" and ``\textit{stacking}" in CNN architecture (subsection \ref{label:cnn}), the authors constructed a deep model by using the Independent Subspace Analysis (ISA) \cite{hyvarinen2009independent} (see Figure~\ref{fig16}a) and Principal Component Analysis (PCA) \cite{mudrova2005principal}. The ISA is trained on small input patches for learning feature directly from unlabeled video data. It is then convolved with a larger region of the input image. The PCA algorithm is applied on the top of ISA  for reducing dimensions. The responses are then used as the input layer for another ISA.

The method is evaluated on KTH \cite{schuldt2004recognizing} , Hollywood2 \cite{marszalek2009actions}, UCF sports\cite{4587727} and YouTube datasets \cite{liu2009recognizing}. Table~\ref{tab5} shows that this deep architecture advanced the state-of-the-art in human action recognition when the paper was published.
\begin{table}[h!]
  \centering
  \caption{\textbf{Comparison of Le's method and the best methods before}}
  \label{tab5}
  \begin{tabular}{ccccc}
  \hline
  \textbf{Method } & \textbf{KTH} & \textbf{Hollyhood2 } & \textbf{UCF} & \textbf{YouTube} \\ 
   \hline
   Measure & AA & Mean AP & AA & AA \\
   \hline
   Le et al. \cite{Le:2011:LHI:2191740.2192108} & \textbf{93.9}\% & \textbf{53.3}\% & \textbf{86.5}\% & \textbf{75.8}\% \\
   Previous best result & 92.1\% & 50.9\% & 85.6\% & 71.2\% \\
   \hline
   Improvements & 1.8\%  & 2.4\%  & 0.9\%  & \textbf{4.6\%} \\
  \end{tabular}
  \begin{flushright}
  {\footnotesize Here, the average accuracy is noted by AA.}
  \end{flushright}
\end{table}

Srivastava \textit{et al}. \cite{srivastava2015unsupervised} constructed a model which consists of two LSTMs -  the encoder LSTM and the decoder LSTM to learn representations of sequences of images. The state of the LSTM encoder is the representation of the input video. Then, the LSTM decoder will reconstruct the input sequence from this representation. It can be used for reconstructing the input sequence as well as predicting the future sequence.

\begin{figure*}
\begin{center}
\begin{tabular}{c}
\includegraphics[width=16cm,height=5cm]{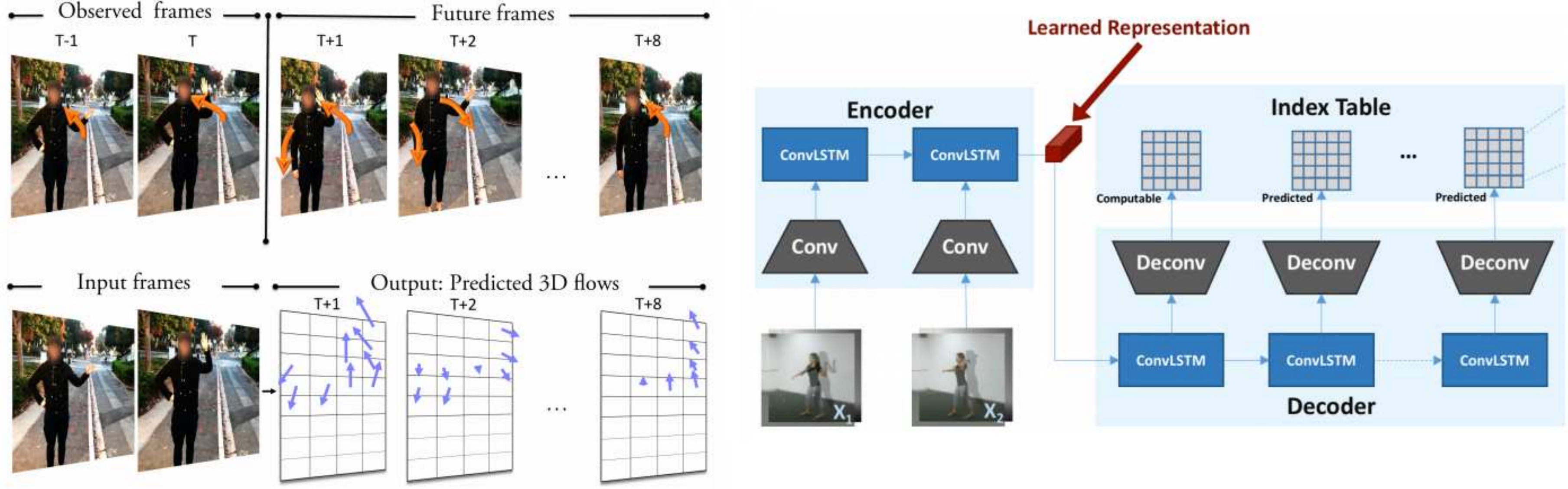}\\ 
\end{tabular}
\end{center}
\caption 
{ \label{figlstm-cnn-autoencoder} \textbf{(a)}  Illustration of the idea of learning a video representation by predicting a sequence of basic motions described as atomic 3D
flows. The learned representation is then used for action recognition. \textbf{(b)} The learning framework architecture based on the Recurrent Neural Network based Encoder-Decoder proposed in the work of Luo \textit{et al.} \cite{Luo2017UnsupervisedLO} }
\end{figure*}

Very recently, Luo \textit{et al.} \cite{Luo2017UnsupervisedLO} combined many different models to build a deep learning framework for recognition human motion in Videos. The idea is designing a network which is able to predict the future 3D motions in videos (see Figure~\ref{figlstm-cnn-autoencoder}). Given input frames, the model will predict 3D flows in future frames, then use the features to recognize activities. To do that, a Recurrent Neural Network based Encoder-Decoder framework has proposed. During the encoding process, CNNs (the standard VGG-16 networks) are used for extracting a low-dimensionality feature from the input frames. Then, the LTSMs have been used to learn the temporal representation of motion. The learned representation is then decoded in the decoding process to generate  the atomic 3D flows. This approach achieved the state-of-the-art result on NTU-RGB+D dataset \cite{DBLP:journals/corr/ShahroudyLNW16} and MSR Daily Activity3D \cite{Li2010ActionRB}. To the best of our knowledge, this model is the best learning framework at the moment for action recognition using different input modalities (RGB, Depth, RGB-D).

A new unsupervised learning approach called Generative Adversarial Networks (GANs) was proposed by Ian \textit{et al.} \cite{Goodfellow2014GenerativeAN}. In 2016, Radford \textit{et al.} \cite{Radford2015UnsupervisedRL} introduced a set of architectures called Deep Convolutional GANs (DCGANs) in order to train GANs in a better way. This study showed that GANs can learn good representations of images for supervised learning and generative modeling. After that, GANs have started to show their real potential. E.g. Vondrick \textit{et al.} \cite{Vondrick2016GeneratingVW} capitalized on recent advances in GANs for both action classification and prediction in video. A two-stream generative model has built for learning  scene dynamics. This study is an open research opportunity for designing of predictive models for understanding human actions.

\section{Discussion}
\label{section:5}
Human action recognition has become one of the most active research topics in computer vision during the last two decades. In particular, the appearance of the DL models as well as the advances of parallel computing techniques, e.g. GPU computing, opened up more new opportunities for this field. Many DL based approaches have developed and applied for various applications related to human action recognition. Their studies indicate various methods to learn motion features from videos and use them to recognize and classify actions. In this section, we provide a detailed analysis of the mentioned classes of architectures. The pros and cons of each class and the link between them will be discussed. Based on these analyses, we point out challenges, current trends and potential directions future research in this field.

After reviewing more than two hundred papers, our study shows that human action recognition has advanced rapidly from recognition in controlled environment with small size benchmark datasets to recognition of actions in realistic videos with very large scale benchmarks. DL techniques play an important role in this progress. In the literature of human action recognition based on DL, CNNs seem to be the most important model for learning spatio-temporal features of human action directly from RGB and RGB-D videos without pre-processing. Almost outstanding architectures, such as networks proposed by Ji \textit{et al}. \cite{ji20133d}, Tran \textit{et al}. \cite{tran2015learning}, Simonyan \textit{et al}. \cite{simonyan2014two}, Wang \textit{et al.} \cite{Wang_UntrimmedNets}, Feichtenhofer \textit{et al.} \cite{feichtenhofer2016spatiotemporal}, Luo \textit{et al.} \cite{Luo2017UnsupervisedLO}, etc. have used 3D convolutional filters to extract motion features. The key ideas behind CNNs allow them to work directly on image structure and obtaining high-level features by composing lower-level ones. CNNs are not only working as an end-to-end solution, they were also used as a feature extractor and were a part in another frameworks. However, CNNs achieve very goof performance when they were trained on very large datasets. If not, overfit will happen. Some techniques have been developed to prevent overfitting in convolutional layers such as dropout, data augmentation (e.g. random cropping, flipping, color effect, etc). When training a very deep CNN architecture, millions of connections between neurons will be involved. Therefore, another limiting factor of CNNs is the high energy consumption due to its high computational complexity. Normally, GPU computing is required to work with this type algorithm.

Recurrent neural networks with long short-term memory (LSTM-RNNs) have been designed for solving time series problems. LSTM-RNNs have been used successfully in modeling the long-term context information of motion sequences, specifically with skeleton data as the work of Du \textit{et al.}\cite{du2015hierarchical}, Song \textit{et al.}\cite{Song2016AnES}, Zhu \textit{et al.} \cite{Zhu2016CooccurrenceFL}, Li \textit{et al.} \cite{Li2016OnlineHA}, Liu \textit{et al.}  \cite{Liu2016SpatioTemporalLW}. The success of LSTM-RNNs for human action recognition comes from their ability to take advantage the entire history motion frames. Even so, most of LSTM-RNN based models can not work directly on raw data. For example, skeleton data need to be preprocessed before feeding into LSTM-RNNs. It is difficult to build an LSTM-RNN based end-to-end learning framework with RGB-D data. Consequently, many authors used CNN to extract color features and then fed into the LSTM for sequences learning and prediction.

Deep belief network (DBNs) and Stacked Denoising Autoencoders (SDAs) are also very promising choice for action recognition tasks. For DBNs, these networks can be trained in an semi-supervised way with less labeled data from a set of examples to classify its inputs. The limitation of DBNs is that they require hand-crafted features \cite{foggia2014exploiting}  or converting input data to appropriate form \cite{ali2014learning}. SDAs can learn motion features in unsupervised manner and are capable of generating robust features. However, it has several drawbacks related to its optimization process.

\subsection{A quantitative analysis}

\begin{flushleft}
$\bullet$ \hspace{0.3cm} \textit{Hand-crafted approaches and deep learning approaches: A small comparison}
\end{flushleft}

In order to have a general view on recognition accuracies reported by hand-crafted approaches and deep learning approaches, we have carried out a small performance comparison on KTH \cite{schuldt2004recognizing} dataset. This dataset has been used to evaluate many action recognition solutions, both the traditional approaches based on hand-crafted features and deep learning based approaches over many years. 

\begin{figure*}
\begin{center}
\begin{tabular}{c}
\includegraphics[width=18cm,height=9.5cm]{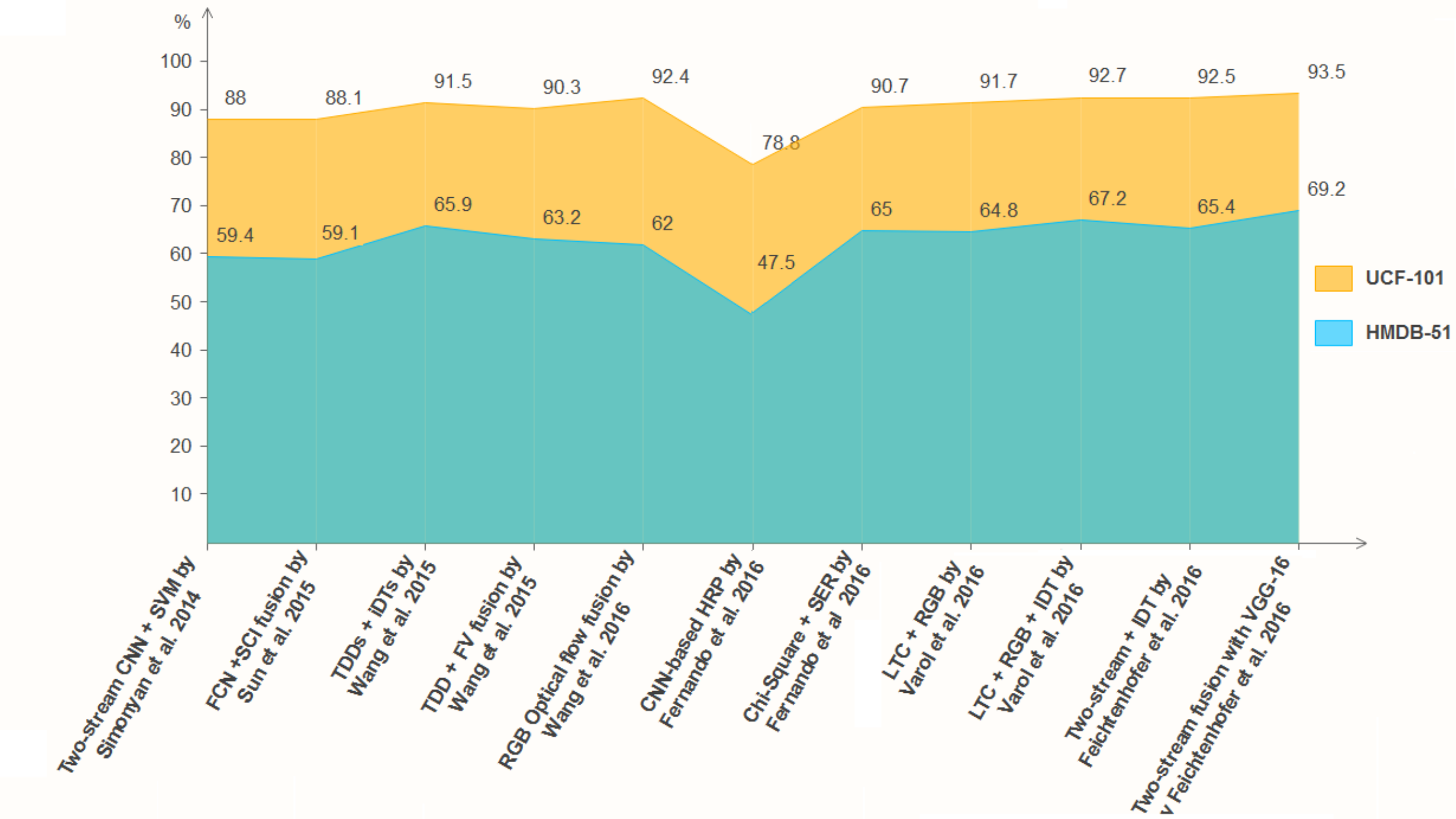}\\ 
\end{tabular}
\end{center}
\caption 
{ \label{fig-dl-comparison} The recognition performance of different deep learning based solutions on HMDB-51 and UCF-101 datasets.}
\end{figure*}

\begin{flushleft}
$\bullet$ \hspace{0.3cm} \textit{A performance comparison between deep learning models}
\end{flushleft}
We provide a quantitative analysis of the deep learning approaches on a state-of-the-art benchmark for human action recognition in realistic and challenging settings. Figure~\ref{fig-dl-comparison} shows our comparison based on the performance of many deep learning solution on UCF-101 dataset that have been reviewed in this paper. This comparison helps us to see clearly the current state of this field and also provide the best architectures proposed in the literature. The accuracies are reported directly from the original papers and all of these work use the same measure. We found that the networks proposed by Varol \textit{et al.} \cite{Varol2016LongtermTC}, Feichtenhofer \textit{et al.} \cite{Feichtenhofer2016ConvolutionalTN}, Tran \textit{et al}. \cite{tran2015learning}.

\subsection{The future of DL for human action recognition}

\begin{flushleft}
$\bullet$ \hspace{0.3cm} \textit{Developing unsupervised learning models}
\end{flushleft}
As labeling of data is very costly in terms of money and manpower, we expect that learning features directly from videos in an unsupervised manner is a very important research direction \cite{lecun2015deep}. Unsupervised learning procedures such as DBNs or deep autoencoders will continue to be developed strongly because they could learn features without requiring labeled data or requiring very limited labeled data in pre-training process.
\begin{flushleft}
$\bullet$ \hspace{0.3cm} \textit{Deeper CNNs}
\end{flushleft}
The success of some very deep learning models such as VGGNet \cite{Simonyan2014VeryDC}, GoogLeNet \cite{Szegedy2015GoingDW}, and ResNets \cite{he2015deep} provided that deeper CNN models can boost the recognition accuracy. It appears that the new algorithms allow us to train deeper network easier. For example, He \textit{et al.} \cite{he2015deep} released ResNets in which it has fewer filters and lower complexity than VGGNet \cite{Simonyan2014VeryDC}. Therefore, we expect deeper CNNs will be more  fully exploited in this field.
\begin{flushleft}
$\bullet$ \hspace{0.3cm} \textit{Combining different deep learning models}
\end{flushleft}
Taking full advantage of the different deep learning models and combining them into a single learning framework is a trend in action recognition. Specifically, the use of CNNs with LSTM-RNNs has improved the stare-of-the-art in many benchmark datasets \cite{Baccouche2011SequentialDL},\cite{Ng2015BeyondSS},\cite{Donahue2015LongtermRC},\cite{Giel2015RecurrentNN},\cite{Sharma2015ActionRU},\cite{Ibrahim2016AHD},\cite{Singh2016AMB},\cite{Li2016ActionRB},\cite{Wu2016ActionRW},\cite{Wang2016HierarchicalAN},\cite{7868164}. We believe that this trend will be continued in the future.
\begin{flushleft}
$\bullet$ \hspace{0.3cm} \textit{Fusion of hand-crated and deep learning solutions}
\end{flushleft}
We found that hand-crafted features such as the trajectory descriptors or optical flow frames have been used in most of state-of-the-art DL models as reported in the work of Varol \textit{et al.} \cite{Varol2016LongtermTC}, Feichtenhofer \textit{et al.} \cite{Feichtenhofer2016ConvolutionalTN}, Tran \textit{et al}. \cite{tran2015learning}, and Wang \textit{et al.} \cite{Wang2016ActionsT}. We expect much of future progress in human action recognition to come from systems that use both hand-crafted and DL solutions to solve challenges in this field.
\begin{flushleft}
$\bullet$ \hspace{0.3cm} \textit{Transfer learning}
\end{flushleft}
One of the main difficulty in training deep networks comes from the scarcity of data. To solve this problem, many authors explored a technique called ''\textit{transfer learning}''. Instead of training an entire deep network from scratch, we pretrain the network on a very large dataset, and then use the network either as an initialization for the task of interest. We believe that this trend will be continued in computer vision, including the human action recognition in video.

\section{Conclusion} \label{section:6}

Our goal in carrying out this research is to bring readers a detailed view of the development process and especially of current progress of deep learning models applied to recognize human action in video. A comprehensive review of various DL architectures and their applications in action recognition and related tasks has been provided over more than two hundred related publications. Our analysis and comparisons about the recognition accuracy between DL based approaches and other techniques shown that deep learning is at the moment the best choice for recognizing and classifying human action as well as predicting human behavior. In addition, the characteristics of the most important DL architectures for action recognition have been also analyzed to provide current trends and open problems for future works in this field. With a list of datasets in different complexity levels, this paper will help interested readers in choosing approximate algorithms and datasets to develop new solutions. Although there has been significant progress over the last years, there are still many challenges in applying DL models to build vision-based action recognition systems and to bring their benefits to our life. We are still looking forward to new DL based approaches to improve the performance of recognition systems while decreasing computational cost and requiring less labeled data. We hope this survey is helpful for researchers in this field.

\ifCLASSOPTIONcompsoc
  
  \section*{Acknowledgments}
\else
  
  \section*{Acknowledgment}
\fi

This work was supported by the Centre d'Etudes et d'Expertise sur les Risques, l'environnement la mobilit\'e et l'am\'enagement (CEREMA). The authors would like to express our thanks to all the people who have made helpful comments and suggestions on a previous draft.

\bibliography{reference}  
\bibliographystyle{IEEEtran}

\begin{IEEEbiographynophoto}{Hieu H. Pham} is a Teaching Fellow at the College of Engineering and Computer Science (CECS), VinUniversity, and a Research Fellow at VinUni-Illinois Smart Health Center. He received his Ph.D. in Computer Science from the Toulouse Computer Science Research Institute (IRIT), University of Toulouse, France, in 2019. Previously, he earned the Degree of Engineer in Industrial Informatics from Hanoi University of Science and Technology (HUST), Vietnam, in 2016. His research interests include Computer Vision, Machine Learning, Medical Image Analysis, and their applications in Smart Healthcare. He is the author, co-author of 30 scientific articles appeared in about 20 conferences and journals such as Computer Vision and Image Understanding, Neurocomputing, International Conference on Medical Image Computing and Computer-Assisted Intervention (MICCAI), Medical Imaging with Deep Learning (MIDL), IEEE International Conference on Image Processing (ICIP), and IEEE International Conference on Computer Vision (ICCV). He is also currently serving as Reviewers for MICCAI, ICCV, CVPR, IET Computer Vision Journal (IET-CVI), IEEE Journal of Biomedical and Health Informatics (JBHI), and Nature Scientific Reports. Before joining VinUniversity, Dr. Hieu worked at Vingroup Big Data Institute (VinBigData) as a Research Scientist and Head of the Fundamental Research Team. With this position, he led several research projects on Medical AI, including collecting various types of medical data, managing and annotating data, and developing new AI solutions for medical analysis.
\\
\\
\textbf{Louahdi Khoudour}, Director of Research, received his Ph.D. in Computer Vision from the University of Lille in 1997 and Habilitation \`a Diriger des Recherches (HDR) degree from the University of Paris in 2006. He worked at Ifsttar (formerly INRETS: French National Institute on traffic and safety research) for many years. He moved to CEREMA in 2011, where he is head of a research group working on safety and security in transport.\\
\\
\textbf{Alain Crouzil} received his Ph.D. degree in Computer science from the Paul Sabatier University of Toulouse in 1997. He is currently an associate professor and a member of the Traitement et Comprehension d'Images group of Institut de Recherche en Informatique de Toulouse. His research interests concern stereo vision, shape from shading, camera calibration, image segmentation, change detection, and motion analysis.\\
\\
\textbf{Pablo Zegers} received his B.S. and P.E. degrees in Engineering from the Pontificia Universidad Catolica, Chile, in 1992, his M.Sc. from The University of Arizona, USA, in 1998, and his Ph.D., also from The University of Arizona, in 2002. He is currently an Associate Professor of the College of Engineering and Applied Sciences of the Universidad de los Andes, Chile. His interests are artificial intelligence, machine learning, neural networks, and information theory. From 2006 to 2010 he was the Academic Director of this College, and for a brief period at the end of 2010, the Interim Dean. He is a Senior Member of the IEEE, and currently the Secretary of the Chilean IEEE section.\\
\\
\textbf{Sergio A. Velastin} obtained his Ph.D. in 1982 from the University of Manchester, UK. He is a professor of applied computer vision and was the director of the Digital Imaging Research Centre at Kingston University until 2012. He then worked as a research professor at the University of Santiago (Chile) and is currently UC3M-Marie Curie Research Professor at the University Carlos III de Madrid, Spain, working on human action recognition. He is a fellow of the IET and a Senior Member of the IEEE. \end{IEEEbiographynophoto}
\end{document}